\documentclass{article}

\usepackage[preprint]{neurips_2026}

\usepackage[utf8]{inputenc}
\usepackage[T1]{fontenc}    
\usepackage{hyperref}      
\usepackage{url}           
\usepackage{booktabs}       
\usepackage{amsmath, amssymb, amsfonts, amsthm}
\usepackage{nicefrac}       
\usepackage{microtype}      
\usepackage{xcolor}         
\usepackage{graphicx}

\usepackage{subcaption} 
\usepackage{multirow}
\usepackage{array}
\usepackage{makecell}
\title{Spatial Lifting for Dense Prediction}

\author{
Mingzhi Xu$^{1}$ \and Tao Zhou$^{1}$ \and Yong Li$^{2}$ \and Yizhe Zhang$^{1}$\thanks{Correspondence to: zhangyizhe@njust.edu.cn} \\
$^{1}$Nanjing University of Science and Technology \\
$^{2}$Southeast University
}

\begin{document}

\maketitle

\begin{abstract}
    We present Spatial Lifting (SL), a novel methodology for dense prediction tasks. SL operates by lifting standard inputs, such as 2D images, into a higher-dimensional space and subsequently processing them using networks designed for that higher dimension, such as a 3D U-Net. Counterintuitively, this dimensionality lifting allows us to achieve good performance on benchmark tasks compared to conventional approaches, while reducing inference costs and \textbf{drastically lowering the number of model parameters}. The SL framework produces intrinsically structured outputs along the lifted dimension. This emergent structure facilitates dense supervision during training and enables single-forward-pass self-consistency-based quality and uncertainty estimation at test time.  Spatial Lifting introduces a simple and general modeling strategy that offers a promising path toward more efficient, accurate, and reliable deep networks for dense prediction tasks in vision. 
    % Code is available at \url{https://anonymous.4open.science/r/Slifting-C575}.
\end{abstract}

\section{Introduction}
Dense prediction tasks, such as semantic segmentation \cite{Ronneberger2015_UNet, Chen2017_DeepLab}, and depth estimation \cite{ranftl2020towards, Bhat2021_AdaBins}, are fundamental to computer vision and widely used in applications like autonomous driving, robotics, and medical imaging. These tasks produce structured pixel-wise or voxel-wise outputs, where accuracy, efficiency, and robustness are critical. Recent deep learning methods, from CNNs \cite{He2016_ResNet, Long2015_FCN} to transformer-based models \cite{Dosovitskiy2021_ViT, Xie2021_SegFormer}, have achieved strong performance but are often computationally expensive, limiting deployment on resource-constrained devices. In addition, uncertainty estimation typically relies on techniques such as ensembling~\cite{lakshminarayanan2017simple} or Monte Carlo dropout~\cite{Gal2016_DropoutUncertainty}, further increasing inference overhead.

% Dense prediction tasks, such as semantic segmentation \cite{Ronneberger2015_UNet, Chen2017_DeepLab}, depth estimation \cite{ranftl2020towards, Bhat2021_AdaBins}, and optical flow \cite{Teed2020_RAFT}, form a cornerstone of computer vision with critical applications ranging from autonomous driving and robotics to medical image analysis and augmented reality. These tasks aim to produce structured outputs by assigning a label or value to each pixel or voxel in the input, where accuracy, efficiency, and robustness are critical for reliable real-world deployment. Significant progress in deep learning has led to highly effective dense prediction models, ranging from convolutional neural networks (CNNs) \cite{He2016_ResNet, Long2015_FCN} to more recent transformer-based architectures \cite{Dosovitskiy2021_ViT, Xie2021_SegFormer}. Despite their strong performance, these models are often computationally expensive due to their large number of parameters and high floating-point operation costs, posing challenges for deployment on resource-constrained platforms like mobile devices or embedded systems. Furthermore, obtaining uncertainty estimation typically requires additional mechanisms such as model ensembling\cite{lakshminarayanan2017simple} or Monte Carlo dropout \cite{Gal2016_DropoutUncertainty}, further increasing inference-time overhead.

\begin{figure}[t]
    \centering
    \includegraphics[width=0.85\textwidth]{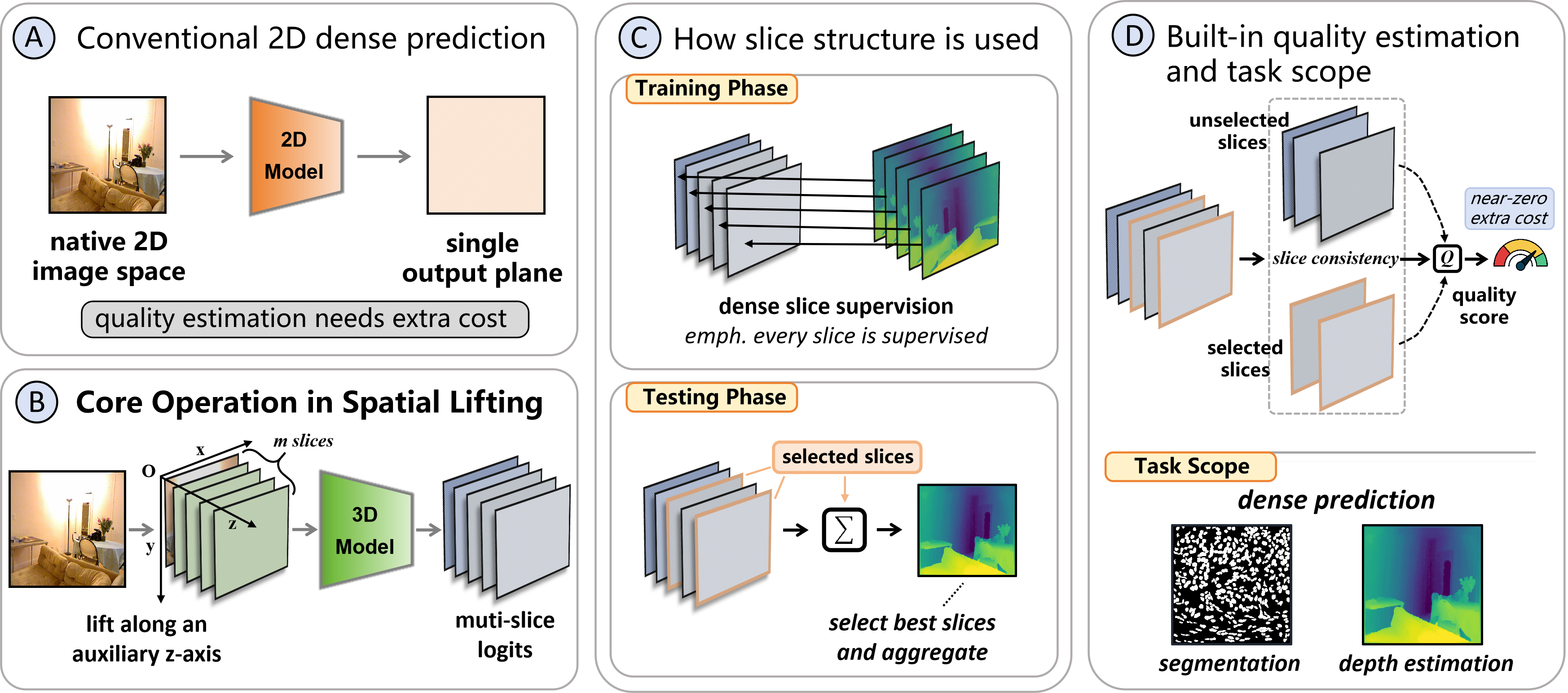}
    \caption{Overview of the proposed Spatial Lifting for dense prediction tasks.}
    \label{fig:overview}
\end{figure}

Most efforts to reduce the computational cost of deep learning, such as efficient architecture design (e.g., \cite{howard2017mobilenets, zhang2018shufflenet, tan2019efficientnet}), post-training pruning (e.g., \cite{han2015learning}), quantization (e.g., \cite{krishnamoorthi2018quantizing}), and knowledge distillation (e.g., \cite{touvron2021training}), operate directly in the native input dimension. However, these approaches largely focus on reducing redundancy within the original spatial representation, without exploring whether computation in a higher-dimensional space could provide additional representational benefits. In this work, we present \textbf{Spatial Lifting (SL)}, an effective and efficient methodology for dense prediction. The core idea is to lift standard inputs, such as 2D images, into a higher-dimensional space (e.g., 3D), where they are processed by neural networks designed for that domain, such as a 3D U-Net \cite{Cicek2016_3DUNet}. We demonstrate that this dimensionality expansion improves performance on established benchmark tasks compared to conventional architectures operating in the original input space. Crucially, this performance gain is achieved while reducing both the number of model parameters and the GMACs required for inference. Beyond efficiency and accuracy, SL intrinsically produces structured representations along the auxiliary lifted dimension without extra computation. This emergent structure enables dense supervision during training and supports efficient quality estimation at test time with near-zero additional overhead. A high-level overview of SL is illustrated in Fig.~\ref{fig:overview}. Our key contributions are summarized as follows:
\textbf{(1) We introduce Spatial Lifting (SL)}, a novel paradigm for dense prediction that lifts 2D inputs into a higher-dimensional space (e.g., 3D) and processes them using higher-dimensional networks, enabling richer spatial representations while reducing computational cost.
\textbf{(2) SL reduces model complexity}, achieving competitive or superior performance with significantly fewer parameters and GMACs compared to conventional decoders.
\textbf{(3) We propose a built-in quality estimation mechanism} that leverages consistency across output slices to estimate segmentation quality and uncertainty with negligible overhead.
\textbf{(4) We provide a theoretical analysis} that interprets SL as a finite-axis, consensus-regularized self-ensemble, explaining how boundary-conditioned slice diversity, dense slice supervision, and aggregation together yield efficient prediction and low-cost self-consistency estimation.
\textbf{(5) Extensive experiments} on 13 semantic segmentation and 6 depth estimation datasets demonstrate the effectiveness, efficiency, and generalizability of SL across tasks and networks.

\section{Related Work}
Our proposed Spatial Lifting (SL) methodology intersects with several research areas, including efficient deep learning architectures, techniques for enhancing spatial reasoning in dense prediction, methods leveraging dimensionality changes, and quality and uncertainty estimation. We position SL relative to these domains below.

\subsection{Dense Prediction Architectures}
Modern dense prediction is primarily based on deep neural networks, particularly Fully Convolutional Networks (FCNs) \cite{Long2015_FCN} and U-Net architectures \cite{Ronneberger2015_UNet, Cicek2016_3DUNet}, which adopt encoder-decoder structures for multi-scale representation learning.  Subsequent work, such as DeepLab variants \cite{Chen2017_DeepLab}, introduced atrous (dilated) convolutions and spatial pyramid pooling \cite{Zhao2017_PSPNet, He2015_SPPNet} to enlarge receptive fields without sacrificing resolution. More recently, Vision Transformers (ViTs) \cite{Dosovitskiy2021_ViT} and their derivatives (e.g., SegFormer \cite{Xie2021_SegFormer}, SETR \cite{Zheng2021_SETR}) have shown strong performance by leveraging global self-attention. Despite these advances, improved accuracy is often achieved through increased architectural complexity, leading to substantial computational cost. SL offers a different path: instead of adding complexity within the native input dimension, it lifts the image to a higher dimension, aiming to capture complex spatial relationships more efficiently within that space using standard operations (e.g., standard 3D convolutions).

\subsection{Efficient Deep Learning}
Addressing the significant computational cost of modern deep learning models has been a major focus of research. Efforts can be broadly grouped into two main areas: designing efficient network architectures from the ground up and compressing large, pre-existing models.

A primary strategy involves designing inherently efficient architectures. Representative methods include MobileNets \cite{howard2017mobilenets}, using depthwise separable convolutions to reduce FLOPs, and ShuffleNets \cite{zhang2018shufflenet}, which employ pointwise group convolutions and channel shuffling for further efficiency. EfficientNet \cite{tan2019efficientnet} further introduces compound scaling to balance network depth, width, and resolution. These methods mainly optimize 2D convolutions or layer combinations within the original data space.

Another line of research focuses on model compression and acceleration. Network pruning \cite{han2015learning, frankle2018lottery} removes redundant parameters in either unstructured or structured forms. Quantization reduces numerical precision of weights and activations (e.g., 32-bit to 8-bit), shrinking model size and improving inference efficiency. Knowledge distillation \cite{hinton2015distilling} transfers knowledge from a large teacher model to a smaller student, improving its performance. Neural architecture search (NAS) \cite{liu2018darts, tan2019mnasnet, cai2018proxylessnas} further automates the design of efficient architectures tailored to hardware and tasks.

SL differs from these approaches by changing the dimensionality of the computation space. Instead of optimizing 2D operations, it performs standard convolutions in a higher-dimensional representation, where spatial relationships can be more explicitly encoded. This reduces dependence on complex 2D filter designs and channel-wise manipulations.

\subsection{Dimensionality Manipulation}
\textbf{Depth-to-Space / Space-to-Depth (Pixel Shuffle):} Depth-to-Space (D2S), also known as Pixel Shuffle \cite{shi2016real}, rearranges channels into spatial dimensions and is commonly used for efficient upsampling in super-resolution. Its inverse performs downsampling via the opposite transformation. D2S reshapes a tensor $(H, W, C \cdot r^2)$ to $(H \cdot r, W \cdot r, C)$. Unlike SL, D2S does not introduce additional feature extraction in the higher-dimensional space, but instead only redistributes information encoded by previous layers.

\textbf{Lifting in Computational Geometry:} SL draws inspiration from ``lifting'' techniques in computational geometry \cite{de2000computational},
for example, lifting 2D points onto a 3D paraboloid transforms Voronoi diagram computation into finding a 3D lower envelope. Such techniques typically map problems to abstract higher-dimensional spaces where geometric structures become simpler or computationally easier to handle. While sharing the principle of increasing dimensionality, SL differs significantly: it aims to enrich spatial feature representations within a learning framework rather than simplify geometric computation, and its lifted space remains a spatial domain where standard convolutional operations are applied instead of an abstract geometric construction.

\subsection{Prediction Quality and Uncertainty Estimation}
Reliable quality and uncertainty estimation is essential for deploying models in safety-critical settings. Standard approaches include Monte Carlo dropout \cite{Gal2016_DropoutUncertainty}, ensembles~\cite{lakshminarayanan2017simple}, Bayesian neural networks\cite{blundell2015weight, gal2016dropout}, test-time augmentation\cite{shanmugam2021better}, or dedicated uncertainty heads\cite{kendall2017uncertainties}, all of which incur additional inference cost. SL provides an alternative by leveraging structure in the lifted dimension. Variations or consistency along this auxiliary axis serve as a proxy for prediction quality and uncertainty, enabling efficient estimation without multiple forward passes or auxiliary modules. In contrast to prior methods, SL uses dimensional lifting not for geometric simplification or data reparameterization, but as a mechanism for efficient spatial feature learning and quality and uncertainty estimation within a unified framework.

\section{Methodology}
\subsection{Method Pipeline}
\noindent\textbf{Input.} The proposed SL technique applies to both image inputs and feature representations, such as those produced by a pretrained transformer-based encoder. We use image inputs for illustration in the following description.

Given a $k$-D image $\mathbf{I}$ (e.g., $\mathbf{I} \in \mathbb{R}^{w \times h \times 3}$ for $k=2$), we lift it to a $(k+1)$-D representation by replicating it $m$ times along a new axis ($m=16$). The lifted tensor $\mathbf{I}' \in \mathbb{R}^{w \times h \times m \times 3}$ is then fed into the subsequent network. This operation is defined as:
\begin{equation}
\mathbf{I}'(x, y, z, c) 
= \mathbf{I}(x, y, c), 
\quad \forall z \in \{1, 2, \dots, m\},
\label{eq:example}
\end{equation}
where $x \in \{1, \dots, w\}$, $y \in \{1, \dots, h\}$, and $c$ denotes the channel dimension. 

\noindent\textbf{Forward Pass.} The lifted $(k+1)$-D image $\mathbf{I}'$ is fed into a $(k+1)$-D network $f_\theta$, where $\theta$ denotes its parameters. For a 2D input, the lifted representation is treated as a 3D volume and processed by a 3D UNet. The network produces a prediction tensor $\mathbf{P} \in \mathbb{R}^{w \times h \times m \times C_{out}}$, where $C_{out}$ is the number of classes: $\mathbf{P} = f_\theta(\mathbf{I}')$. The output, logits maps $\mathbf{P}$, contains $m$ slices along the lifted dimension.

\noindent\textbf{Model Training with Dense Slice Supervision.} Given a $k$-D training set $\{(\mathbf{I}^{(i)}, \mathbf{M}^{(i)})\}_{i=1}^n$, each ground-truth mask is replicated along the lifted dimension to form:
\begin{equation} \mathbf{M}'^{(i)}(x,y,z,c) = \mathbf{M}^{(i)}(x,y,c), \quad z \in \{1,\dots,m\}. \end{equation}

The network prediction $\mathbf{P}^{(i)} = f_\theta(\mathbf{I}'^{(i)})$ is supervised with a slice-wise loss $\ell(\cdot,\cdot)$, such as Dice loss or cross-entropy:
\begin{equation}
\mathcal{L}_{\text{train}} =
\frac{1}{m} \sum_{z=1}^m
\ell\!\left(\mathbf{P}^{(i)}(\cdot,\cdot,z,\cdot),\mathbf{M}^{(i)}(\cdot,\cdot,\cdot)\right).
\end{equation}

At the end of training, we compute the average loss for each slice $z \in \{1,\dots,m\}$ and select the $s$ slices $\{z_1, z_2, \dots, z_s\}$ with the lowest values:
\begin{equation}
\{z_1, z_2, \dots, z_s\} =
\operatorname{arg\,min}_{S \subset \{1,\dots,m\},\, |S|=s}
\sum_{z \in S}
\ell\!\left(\mathbf{P}^{(i)}(\cdot,\cdot,z,\cdot), \mathbf{M}^{(i)}(\cdot,\cdot,\cdot)\right), \quad z \in \{1,\dots,m\},
\end{equation}
where $z_1, z_2, \dots, z_s$ are indices corresponding to the selected slices. The default value of $s$ is set to 5.

\noindent\textbf{Model Testing.} 
During testing, a $k$-D image $\mathbf{I}_{\text{test}} \in \mathbb{R}^{w \times h \times 3}$ (here $k$ equals 2) is lifted to a 3-D representation $\mathbf{I}'_{\text{test}} \in \mathbb{R}^{w \times h \times m \times 3}$ and passed through the trained network $f_\theta$: $\mathbf{P}_{\text{test}} = f_\theta(\mathbf{I}'_{\text{test}})$.

The final prediction is obtained by averaging the logits over the selected slices $\mathcal{Z}=\{z_1,\dots,z_s\}$, and \( \sigma \) denotes the activation function, e.g., the sigmoid function:
\begin{equation}
\text{Segmentation}(x, y)=\arg\max_c \,\sigma\!\left(\sum_{z\in\mathcal{Z}} \mathbf{P}_{\text{test}}(x,y,z,c)\right).
\end{equation}

\subsection{Quality Estimation via Slice Consistency}
\label{sec:quality_estimation}

SL produces \(m\) slice-wise predictions for a single input. After training, Eq.~(4) selects a subset of
reliable slices \(Z=\{z_1,\dots,z_s\}\); let \(Z^c := \{1,\dots,m\}\setminus Z\) denote the remaining slices.
For segmentation, let \(P_B(\cdot,\cdot,z,\cdot)\) be the binarized prediction at slice \(z\). We define the
slice-consistency score
\begin{equation}
Q
=
\frac{1}{|Z|\,|Z^c|}
\sum_{i\in Z}\sum_{j\in Z^c}
\mathrm{Dice}\!\left(
P_B(\cdot,\cdot,i,\cdot),
P_B(\cdot,\cdot,j,\cdot)
\right).
\label{eq:quality_score}
\end{equation}

A high value of \(Q\) indicates strong agreement between the selected and unselected slices, suggesting
that the internal slice family has concentrated around a stable solution. A low value of \(Q\) indicates
substantial cross-slice disagreement, reflecting ambiguity or instability in the prediction. Therefore,
\(Q\) serves as a deterministic \emph{self-consistency} proxy for prediction quality: larger \(Q\) typically
implies better expected performance.

When an uncertainty score is required, we estimate it from the internal predictions without additional forward passes. 
% Specifically, the quality score \(Q\) is computed as the mean pairwise Dice similarity between slice-level predictions, while uncertainty \(U\) is defined as the variance of these similarities. Unlike MC Dropout or ensembles, both \(Q\)
% and \(U\) are obtained from a single forward pass and incur negligible additional computational cost. As shown in Section~4, \(Q\) correlates with ground-truth Dice, while \(U\) provides a useful ranking signal for risk-coverage evaluation.
Specifically, the quality score \(Q\) is computed as the mean pairwise Dice similarity between slice-level predictions. The primary uncertainty score \(U_{\mathrm{disp}}\) is defined as \(U_{\mathrm{disp}} = 1 - Q\), while the variance of these similarities \(U_{\mathrm{var}}\) is additionally reported to characterize the heterogeneity of cross-slice agreement. Unlike MC Dropout or ensemble-based methods, both \(Q\) and \(U\) (\(U_{\mathrm{disp}}, U_{\mathrm{var}}\)) are obtained from a single forward pass and incur negligible additional computational overhead. As shown in Section~4, \(Q\) exhibits strong correlation with ground-truth Dice, whereas \(U\) provides an effective signal for risk-coverage evaluation and uncertainty-based ranking.

\subsection{Theoretical Analysis}

Spatial Lifting (SL) does not inject new information into the input. The lifted tensor is obtained by replicating the same image along an auxiliary axis, so any gain must come from the inductive bias induced by finite-axis computation and the shared training objective. We therefore interpret SL not as learning in a genuinely richer input space, but as constructing a structured family of coupled predictors that are regularized toward agreement during training and aggregated at test time.

Let $h_z^{(0)} = I$ for $z \in \{1,\dots,m\}$. Consider a network whose interactions along the lifted axis are local:
\begin{equation}
h_z^{(\ell+1)}
=
\psi^{(\ell)}\!\left(
\sum_{\tau=-r_\ell}^{r_\ell}
K_{\tau}^{(\ell)} *_{xy} h_{z-\tau}^{(\ell)}
\right),
\end{equation}
where $K_{\tau}^{(\ell)}$ is the 2D kernel associated with offset $\tau$, $*_{xy}$ denotes 2D convolution in the native spatial dimensions, $\psi^{(\ell)}$ is a pointwise nonlinearity, and out-of-range indices are zero padded.

\paragraph{Boundary-conditioned slice family.}
For each slice $z$, there exists a slice-dependent 2D mapping $g_{\theta,z}$ such that
\begin{equation}
P_z = g_{\theta,z}(I).
\end{equation}
All slice predictors share the same parameters; their differences arise only from finite-axis boundary effects and local coupling along the lifted axis. In particular, if the network were equivariant to cyclic shifts along $z$, then a replicated input would produce identical outputs on all slices. Hence non-trivial slice diversity in SL is not created by additional input information, but by the finite, non-periodic auxiliary axis. For a single lifted convolution applied to a replicated input,
\begin{equation}
Y_z
=
\sum_{\tau=-r}^{r}
\mathbf{1}\{1 \le z-\tau \le m\}\,(K_\tau *_{xy} I)
=
K^{\mathrm{eff}}_z *_{xy} I,
\end{equation}
where
\begin{equation}
K^{\mathrm{eff}}_z
=
\sum_{\tau=-r}^{r}
\mathbf{1}\{1 \le z-\tau \le m\} K_\tau .
\end{equation}
Thus each slice implements a slice-dependent effective 2D operator determined by its distance to the lifted-axis boundaries.

\paragraph{Dense slice supervision as consensus regularization.}
At a fixed spatial location, let $u_1,\dots,u_m \in \mathbb{R}^C$ denote the slice-wise logits and let $y$ be the common target. Assume the per-location loss has the form
\begin{equation}
\ell(y,u)=\Phi(u)-\langle T(y),u\rangle+c(y),
\end{equation}
which includes multiclass cross-entropy, binary logistic loss, and Brier loss. Define the slice average
\begin{equation}
\bar u = \frac{1}{m}\sum_{z=1}^{m} u_z
\end{equation}
and the Bregman divergence
\begin{equation}
D_\Phi(a,b)=\Phi(a)-\Phi(b)-\langle \nabla \Phi(b), a-b\rangle.
\end{equation}
Then
\begin{equation}
\frac{1}{m}\sum_{z=1}^{m}\ell(y,u_z)
=
\ell(y,\bar u)
+
\frac{1}{m}\sum_{z=1}^{m} D_\Phi(u_z,\bar u).
\end{equation}
This identity shows that dense slice supervision simultaneously optimizes the mean predictor and penalizes inter-slice disagreement. Therefore, SL is not merely a multi-output architecture: it is a consensus-regularized slice ensemble. From an optimization viewpoint, this also implies that the parameter update is driven by the average of slice-wise gradients, which dampens slice-specific noise and biases learning toward solutions that perform well across the entire slice family rather than on a single internal path.

\paragraph{Aggregation and self-consistency.}
Let $p_z=\sigma(P_z)$ denote the slice-wise predictions after activation. For a selected slice set $Z$, define the aggregated prediction
$
\hat p_Z = \frac{1}{|Z|}\sum_{z\in Z} p_z .
$
Assume
$
p_z = p^\star + b_z + \varepsilon_z,
\qquad
\mathbb{E}[\varepsilon_z]=0,
$
where $p^\star$ is the desired prediction, $b_z$ is a slice-dependent bias term, and $\varepsilon_z$ is a residual term. Writing
$
\bar b_Z = \frac{1}{|Z|}\sum_{z\in Z} b_z,
\qquad
\Sigma_{ij} = \mathrm{Cov}(\varepsilon_i,\varepsilon_j),
$
the aggregated error satisfies
$
\mathbb{E}\|\hat p_Z-p^\star\|_2^2
=
\|\bar b_Z\|_2^2
+
\mathrm{tr}\!\left(
\frac{1}{|Z|^2}\sum_{i,j\in Z}\Sigma_{ij}
\right).
$
Hence aggregation is beneficial when the selected slices do not share a large common bias and their residual errors are not perfectly correlated.

To connect this to quality estimation, define the cross-set dispersion
$
\Delta_Z
=
\frac{1}{|Z|(m-|Z|)}
\sum_{i\in Z}\sum_{j\notin Z}
\|p_i-p_j\|_2^2 .
$
A direct expansion gives
$
\mathbb{E}[\Delta_Z]
=
\frac{1}{|Z|(m-|Z|)}
\sum_{i\in Z}\sum_{j\notin Z}
\Big(
\|b_i-b_j\|_2^2
+
\mathrm{tr}(\Sigma_{ii}+\Sigma_{jj}-2\Sigma_{ij})
\Big).
$

Thus low cross-slice dispersion indicates both small bias gaps and small residual disagreement across the slice family. The practical score $Q$ in Eq.~(6), which measures agreement between selected and unselected slice predictions, can therefore be interpreted as a bounded, task-specific surrogate for low $\Delta_Z$: larger $Q$ indicates a more concentrated internal slice family and, in expectation, a more reliable aggregated prediction.

\paragraph{Implications.}
This analysis yields a narrower but stronger interpretation of SL. The method does not help because lifting injects new information into the input. Rather, it constructs a low-cost family of coupled predictors whose diversity is induced by the finite auxiliary axis, whose disagreement is explicitly suppressed by dense supervision, and whose residual errors can be reduced by aggregation. Accordingly, SL should be most effective when the slice family exhibits modest diversity with imperfectly correlated residuals; its benefit should diminish when all slices collapse to nearly the same predictor or when their errors are dominated by a shared systematic bias.
\subsection{Parameter and 
Computational Complexity}
\noindent \textbf{Traditional Design:} A 2D convolutional layer $l$ (kernel $K_x, K_y$) has $K_x K_y C_l C_{l+1} + C_{l+1}$ parameters. Total parameters for $L$ intermediate layers:  
$\sum_{l=0}^{L-1} ( K_x K_y C_l C_{l+1} + C_{l+1} ) + ( K_x K_y C_L C_{\text{out}} + C_{\text{out}} )$.  
MACs per layer: $2 K_x K_y w h C_l C_{l+1}$.  
Total MACs for $L$ intermediate layers and one output layer:  
$\sum_{l=0}^{L-1} ( 2 K_x K_y w h C_l C_{l+1} ) + ( 2 K_x K_y w h C_L C_{\text{out}} )$.  

\noindent \textbf{SL-Design:} With a constant number of intermediate channels $C_*$, a $3$-D intermediate layer has $K_x K_y K_z C_*^2 + C_*$ parameters. The final output layer producing $C_{\text{out}}$ classes has $K_x K_y K_z C_* C_{\text{out}} + C_{\text{out}}$ parameters. Total parameters for $L$ intermediate layers:  
$L (K_x K_y K_z C_*^2 + C_*) + (K_x K_y K_z C_* C_{\text{out}} + C_{\text{out}})$.  
MACs per intermediate layer: $2 K_x K_y K_z w h m C_*^2$.  
Total MACs for $L$ intermediate layers and one output layer:  
$L ( 2 K_x K_y K_z w h m C_*^2 ) + ( 2 K_x K_y K_z w h m C_* C_{\text{out}} )$.  

\noindent Compared to conventional architectures where channel width typically increases with depth ($C_l \propto 2^l$), leading to quadratic growth in parameters and MACs, the SL design maintains fixed channel width $C_*$ and avoids such scaling. While computation scales linearly with the lifted dimension $m$, this is controlled via the choice of $C_*$ values (e.g., $C_* = 8$). This results in a more stable computational profile under small $C_*$ settings, with empirical evaluation provided in the experiments section.

% \noindent The SL design when using a small $C_*$ can effectively mitigate the quadratic parameter growth typically observed in traditional architectures employing channel scaling ($C_l \propto 2^l$). While the MACs for SL-design scale linearly with the lifted dimension $m$, conventional designs exhibit substantial MAC increases due to channel scaling. Notably, for small $C_*$ values (e.g., $C_* = 8$), the SL design can achieve comparable or  better predictive accuracy with significantly reduced parameter counts and computational costs, including the lifting process. Empirical results supporting these claims are presented in the subsequent experiments.

\section{Experiments}
Our experiments address two primary objectives. First, we investigate how the SL technique impacts performance and resource consumption (GMACs and model parameters) when applied to widely used architectures such as UNet~\cite{Ronneberger2015_UNet}. Second, we assess whether applying SL on vision decoders for processing features generated by pre-trained encoders (e.g., PVTv2~\cite{wang2022pvt}) can enhance both performance and efficiency. To thoroughly evaluate the proposed approach, our experiments focus on dense prediction tasks, encompassing not only semantic segmentation but also depth estimation, thereby demonstrating the versatility and effectiveness of SL across different applications. All experiments are conducted on a machine equipped with an NVIDIA L20 GPU and 256G RAM.

% \begin{table}[t]
% \footnotesize
% \centering
% \footnotesize
% \caption{MACs and Parameters of UNet without and with SL. Best values are highlighted in \textbf{bold}.}
% \begin{tabular}{@{}c c c c c c@{}}
% \toprule
% \textbf{Metric} & \textbf{Model} & \textbf{5L, 1Res} & \textbf{8L, 1Res} & \textbf{5L, 2Res} & \textbf{8L, 2Res} \\
% \midrule
% \multirow{2}{*}{MACs (G)} 
% & UNet     & 13.7245 & 17.4800 & 18.5698 & 20.7387 \\
% & SL-UNet  & \textbf{\underline{6.0206}} & \textbf{\underline{6.0216}} & \textbf{\underline{7.0719}} & \textbf{\underline{7.0713}} \\
% \midrule
% \multirow{2}{*}{Params (M)}
% & UNet     & 3.3522 & 33.0124 & 6.4959 & 43.2355 \\
% & SL-UNet  & \textbf{\underline{0.0295}} & \textbf{\underline{0.0371}} & \textbf{\underline{0.0382}} & \textbf{\underline{0.0464}} \\
% \bottomrule
% \end{tabular}
% \label{Tab:Metrics}
% \end{table}

\textbf{Dataset}. We conduct experiments on 19 publicly available datasets for semantic segmentation and depth estimation. For semantic segmentation, we use 13 datasets covering multiple anatomical structures (e.g., retina, skin, gastrointestinal tract, glands, and various cell types) and imaging modalities including fundus photography, dermoscopy, histopathology, endoscopy, and microscopy (CHASE\_DB1~\cite{fraz2012ensemble}, DSB2018~\cite{Caicedo2019NucleusSA}, Kvasir\text{-}SEG~\cite{jha2020kvasir}, MoNuSAC~\cite{verma2021monusac2020}, ISIC2018~\cite{codella2018skin,tschandl2018ham10000}, MoNuSeg~\cite{kumar2017dataset}, Sartorius~\cite{sartorius-cell-instance-segmentation}, TNBC~\cite{8438559}, Fluorescent Neuronal Cells~\cite{hitrec2019neural,morelli2021automating}, GlaS~\cite{sirinukunwattana2017gland,sirinukunwattana2015stochastic}, BCCD~\cite{Sarker2023}, FIVES~\cite{jin2022fives}, NuInsSeg~\cite{mahbod2023nuinsseg}). For depth estimation, we employ 6 datasets spanning outdoor and indoor scenes (Cityscapes~\cite{cordts2016cityscapes,liu2019end}, Make3D~\cite{saxena2005learning,saxena2007learning}, DIODE~\cite{vasiljevic2019diode}, KITTI~\cite{geiger2013vision}, NYU Depth~\cite{silberman2012indoor}, MODEST~\cite{shan2021modestkaggle}).

\begin{table}[t]
\footnotesize
\centering
\caption{Comparison of UNet and SL-UNet across datasets, reporting efficiency (MACs and parameters) alongside Dice score (\%). The best results are highlighted in \textbf{bold and underlined}.}
\resizebox{0.82\linewidth}{!}{
\begin{tabular}{@{}l|cc|cc|cc|cc@{}}
\toprule
\multirow{2}{*}{\textbf{Dataset}} 
& \multicolumn{2}{c|}{\textbf{5L,1Res}} 
& \multicolumn{2}{c|}{\textbf{5L,2Res}} 
& \multicolumn{2}{c|}{\textbf{8L,1Res}} 
& \multicolumn{2}{c}{\textbf{8L,2Res}} \\
\cmidrule(lr){2-9}
& UNet & SL-UNet & UNet & SL-UNet & UNet & SL-UNet & UNet & SL-UNet \\
\midrule

DSB2018 
& 87.27 & \textbf{\underline{87.82}}
& 87.76 & \textbf{\underline{90.00}}
& 87.96 & \textbf{\underline{87.97}}
& 87.85 & \textbf{\underline{88.91}} \\

Kvasir-SEG 
& \textbf{\underline{52.62}} & 52.35
& \textbf{\underline{71.73}} & 69.50
& \textbf{\underline{64.68}} & 57.49
& \textbf{\underline{81.16}} & 76.03 \\

ISIC2018
& \textbf{\underline{80.22}} & 78.81
& 75.33 & \textbf{\underline{82.46}}
& 79.58 & \textbf{\underline{85.28}}
& 82.26 & \textbf{\underline{82.97}} \\

Sartorius 
& 61.77 & \textbf{\underline{69.85}}
& 70.92 & \textbf{\underline{70.93}}
& 61.90 & \textbf{\underline{70.80}}
& 70.94 & \textbf{\underline{73.02}} \\

MoNuSAC 
& 62.73 & \textbf{\underline{66.21}}
& 61.99 & \textbf{\underline{66.61}}
& 62.92 & \textbf{\underline{66.76}}
& 69.72 & \textbf{\underline{70.95}} \\

CHASE\_DB1 
& 59.72 & \textbf{\underline{66.14}}
& 60.71 & \textbf{\underline{66.10}}
& 58.87 & \textbf{\underline{66.57}}
& \textbf{\underline{73.19}} & 72.39 \\

MoNuSeg 
& 70.79 & \textbf{\underline{73.20}}
& 70.37 & \textbf{\underline{73.17}}
& 70.83 & \textbf{\underline{72.86}}
& 75.07 & \textbf{\underline{79.02}} \\

TNBC 
& 57.84 & \textbf{\underline{62.93}}
& 63.00 & \textbf{\underline{63.08}}
& 59.77 & \textbf{\underline{62.52}}
& \textbf{\underline{72.02}} & 69.69 \\

FNC 
& 55.89 & \textbf{\underline{65.60}}
& 55.85 & \textbf{\underline{61.47}}
& 57.32 & \textbf{\underline{64.19}}
& 61.35 & \textbf{\underline{68.83}} \\

GlaS 
& 75.27 & \textbf{\underline{76.27}}
& 76.23 & \textbf{\underline{80.49}}
& 77.86 & \textbf{\underline{78.53}}
& 81.65 & \textbf{\underline{84.61}} \\

BCCD 
& 93.71 & \textbf{\underline{95.25}}
& 93.99 & \textbf{\underline{94.36}}
& 93.80 & \textbf{\underline{94.13}}
& 95.05 & \textbf{\underline{95.70}} \\

FIVES 
& 71.82 & \textbf{\underline{73.05}}
& 71.97 & \textbf{\underline{74.57}}
& 70.99 & \textbf{\underline{71.28}}
& 78.59 & \textbf{\underline{79.05}} \\

NuInsSeg 
& 59.50 & \textbf{\underline{60.76}}
& 62.21 & \textbf{\underline{67.54}}
& 62.51 & \textbf{\underline{63.23}}
& 64.59 & \textbf{\underline{65.96}} \\

\midrule
\textbf{MACs (G)} 
& 13.7245 & \textbf{\underline{6.0206}}
& 18.5698 & \textbf{\underline{7.0719}}
& 17.4800 & \textbf{\underline{6.0216}}
& 20.7387 & \textbf{\underline{7.0713}} \\

\textbf{Params (M)} 
& 3.3522 & \textbf{\underline{0.0295}}
& 6.4959 & \textbf{\underline{0.0382}}
& 33.0124 & \textbf{\underline{0.0371}}
& 43.2355 & \textbf{\underline{0.0464}} \\

\bottomrule
\end{tabular}
}
\label{Tab:UNet_SLUNet}
\end{table}

% \begin{table}[t]
% \footnotesize
% \centering
% \caption{Decoder architectures: channel dimensions, depth, and complexity metrics.}

% \resizebox{0.6\linewidth}{!}{
% \begin{tabular}{@{}lcccc@{}}
% \toprule
% \textbf{Decoder} & \textbf{Channel} & \textbf{Depth} & \textbf{MACs (G)$\downarrow$} & \textbf{Params (M)$\downarrow$} \\
% \midrule
% HSNet & [64,128,320,512] & -- & 1.5653 & 4.4072 \\
% SL-UNet(5L, 1Res) & [8,8,8,8,8,8,8,8] & 16 & \textbf{\underline{1.2653}} & \textbf{\underline{0.5414}} \\
% SL-UNet(8L, 1Res) & [8,8,8,8,8,8,8,8] & 16 & 1.2835 & 0.5648 \\
% SL-UNet(5L, 2Res) & [8,8,8,8,8,8,8,8] & 16 & 1.3296 & 0.5762 \\
% SL-UNet(8L, 2Res) & [8,8,8,8,8,8,8,8] & 16 & 1.3499 & 0.6020 \\
% \midrule
% EMCAD & [64,128,320,512] & -- & 1.1318 & 1.9135 \\
% SL-EMCAD & [16,32,64,128] & [8,4,2,1] & \textbf{\underline{1.0467}} & \textbf{\underline{0.2623}} \\
% \midrule
% CASCADE & [64,128,320,512] & -- & 6.8730 & 10.4219 \\
% SL-CASCADE & [16,32,64,128] & [8,4,2,1] & \textbf{\underline{4.7814}} & \textbf{\underline{1.6409}} \\
% \midrule
% FastDepth & [1280,16,24,32] & -- & 0.2311 & 0.1278 \\
% SL-FastDepth & [16,8,8,8] & 16 & \textbf{\underline{0.2248}} & \textbf{\underline{0.0404}} \\
% \midrule
% MiDaS & [256,256,256,256] & -- & 15.4028 & 8.5934 \\
% SL-MiDaS & [16,16,16,16] & 16 & \textbf{\underline{5.3586}} & \textbf{\underline{0.1055}} \\
% \midrule
% FLDecoder & [64,128,256,512] & -- & 10.3408 & 4.0604 \\
% SL-FLDecoder & [32,16,8,8] & 16 & \textbf{\underline{0.7684}} & \textbf{\underline{0.0506}} \\
% \bottomrule
% \end{tabular}
% }
% \label{Tab:arch_details}
% \end{table}

\textbf{Spatial Lifting on UNet.} Table~\ref{Tab:UNet_SLUNet} compares the model parameter count and inference cost (in GMACs) of the original UNet and SL-UNet. The notation ``5L'' and ``1Res'' respectively denote a five-layer UNet and the inclusion of a single residual connection per layer. Both implementations use the Monai package. 
% In the conventional five-layer U-Net architecture, the number of feature channels is typically configured as 32, 64, 128, 256, and 512 from the first to the fifth layer, whereas SL-UNet adopts a fixed channel setting of 8 and lifts the 2D input to 3D with a depth of 16. SL-UNet features an extremely small number of model parameters and lower GMACs, with extensive parameter sharing across the 3D operations, resulting in a highly lightweight design.
Table~\ref{Tab:UNet_SLUNet} presents segmentation performance of UNet and SL-UNet across 13 benchmark datasets. Overall, SL-UNet often outperforms the original UNet, achieving higher Dice scores while substantially reducing computational cost and parameter counts. For example, for the 5L,1Res configuration, SL-UNet reduces MACs by approximately 56\% (from 13.72G to 6.02G) and decreases parameter counts by over 99\% (from 3.35M to 0.0295M), whereas for the larger 8L,2Res configuration, MACs are reduced by roughly 66\% (from 20.74G to 7.07G) and parameters by nearly 99.9\% (from 43.24M to 0.0464M), showing that the efficiency gains are more pronounced for larger models and indicating greater potential for scaling high-capacity architectures. On smaller or simpler datasets (e.g., FNC, MoNuSAC), SL-UNet shows pronounced improvements, while on high-detail datasets (e.g., ISIC, Sartorius), it still matches or slightly exceeds UNet performance. These results indicate that SL-UNet achieves competitive segmentation accuracy with dramatically lower computational and memory overhead, offering a promising approach for lightweight dense prediction tasks.

% Additionally, Table~\ref{Tab:quality estimation} demonstrates the prediction quality assessment of the SL technique. SL-UNet produces predicted Dice scores that exhibit moderate to high correlations with the actual Dice scores across most datasets, further validating its utility for reliable AI deployment. To better understand how different depth slices contribute to the final prediction, visualization results using various interpretability methods are shown in Fig.~\ref{fig:visualizations}. Specifically, Grad-CAM (a), saliency maps (b), and guided backpropagation (c) highlight the informative regions captured by each slice within the SL decoder.

\textbf{Spatial Lifting on Vision Decoders.} 
We adopt a pre-trained PVTv2 encoder and apply SL to enhance decoder efficiency for dense prediction. Table~\ref{tab:SL_on_vision_decoder} presents the performance of different decoders across benchmark datasets, together with their configuration settings, including channel dimensions, depth, MACs, and parameter counts. For performance comparison, we include HSNet’s decoder~\cite{ZHANG2022106173}, the recently proposed state-of-the-art efficient decoder EMCAD~\cite{rahman2024emcad}, and its SL-enhanced variant, SL-EMCAD. The results show that SL framework improves performance across most datasets while reducing parameter counts. For example, SL-EMCAD lowers parameters from 1.91M to 0.26M while achieving higher Dice scores on most datasets. SL-CASCADE similarly reduces MACs and parameters compared to the original CASCADE. These results demonstrate that SL provides a lightweight, effective enhancement applicable to multiple decoder architectures.

\begin{table}[t]
\footnotesize
\centering
\caption{Spatial Lifting for semantic segmentation: performance comparison measured by Dice score.}

\resizebox{\linewidth}{!}{
\begin{tabular}{l|ccc|cc|cc}
\toprule
\textbf{Model} 
& \textbf{HSNet} 
& \textbf{UNet} 
& \textbf{SL-UNet} 
& \textbf{CASCADE} 
& \textbf{SL-CASCADE} 
& \textbf{EMCAD} 
& \textbf{SL-EMCAD} \\
\midrule

\textbf{Channel} 
& [64,128,320,512] 
& \makecell{[32, 64, 128, 256,\\ 512, 512, 512, 512]}
& [8,8,8,8,8,8,8,8] 
& [64,128,320,512] 
& [16,32,64,128] 
& [64,128,320,512] 
& [16,32,64,128] \\

\textbf{MACs (G)$\downarrow$} 
& 1.5653 
& 21.5863
& \textbf{\underline{1.3499}} 
& 6.8730 
& \textbf{\underline{4.7814}} 
& 1.1318 
& \textbf{\underline{1.0467}} \\

\textbf{Params (M)$\downarrow$} 
& 4.4072 
& 37.1747
& \textbf{\underline{0.6020}} 
& 10.4219 
& \textbf{\underline{1.6409}} 
& 1.9135 
& \textbf{\underline{0.2623}} \\

\midrule

DSB2018     & 86.09 & 89.60 & \textbf{\underline{90.43}} & \textbf{\underline{90.48}} & 89.92 & \textbf{\underline{90.56}} & 87.38 \\
Kvasir-SEG  & 68.39 & \textbf{\underline{84.11}} & 81.20 & 65.96 & \textbf{\underline{66.43}} & \textbf{\underline{70.96}} & 64.48 \\
ISIC2018    & 79.27 & 86.28 & \textbf{\underline{87.07}} & 77.32 & \textbf{\underline{78.72}} & 78.41 & \textbf{\underline{79.93}} \\
Sartorius   & 65.28 & \textbf{\underline{70.51}} & 68.57 & 68.29 & \textbf{\underline{72.98}} & 63.26 & \textbf{\underline{66.62}} \\
MoNuSAC     & 59.04 & 63.82 & \textbf{\underline{66.72}} & \textbf{\underline{63.25}} & 62.79 & 59.89 & \textbf{\underline{60.37}} \\
CHASE\_DB1  & 60.92 & 64.06 & \textbf{\underline{67.07}} & 70.58 & \textbf{\underline{71.17}} & 70.12 & \textbf{\underline{70.47}} \\
MoNuSeg     & 69.84 & 73.81 & \textbf{\underline{76.31}} & 75.16 & \textbf{\underline{76.86}} & 75.86 & \textbf{\underline{76.31}} \\
TNBC        & 65.75 & 62.51 & \textbf{\underline{70.72}} & 59.61 & \textbf{\underline{64.83}} & 65.09 & \textbf{\underline{67.97}} \\
FNC         & 55.42 & 56.22 & \textbf{\underline{57.37}} & 51.04 & \textbf{\underline{51.54}} & 60.75 & \textbf{\underline{61.19}} \\
GlaS        & 79.79 & 84.70 & \textbf{\underline{87.05}} & 83.17 & \textbf{\underline{85.08}} & 79.84 & \textbf{\underline{84.13}} \\
BCCD        & 92.35 & 94.64 & \textbf{\underline{95.63}} & \textbf{\underline{95.82}} & 94.89 & \textbf{\underline{95.36}} & 95.14 \\
FIVES       & 71.03 & 76.36 & \textbf{\underline{76.69}} & \textbf{\underline{80.24}} & 77.56 & \textbf{\underline{80.30}} & 77.72 \\
NuInsSeg    & 65.94 & \textbf{\underline{66.62}} & 65.34 & 69.08 & \textbf{\underline{69.99}} & \textbf{\underline{70.34}} & 67.51 \\
\bottomrule
\end{tabular}%
}
\label{tab:SL_on_vision_decoder}
\end{table}

\begin{table}[t]
\footnotesize
\centering
\caption{Spatial Lifting for depth estimation: performance comparison across multiple datasets, with corresponding architectural complexity details.}

\resizebox{\linewidth}{!}{
\begin{tabular}{ll|cc|cc|cc}
\toprule
& \textbf{Model} 
& \textbf{\makecell{MobileNet+\\FastDepth}} 
& \textbf{\makecell{MobileNet+\\SL-FastDepth}} 
& \textbf{\makecell{ResNeXt+\\MiDaS}} 
& \textbf{\makecell{ResNeXt+\\SL-MiDaS}} 
& \textbf{\makecell{FLEncoder\\FLDecoder}} 
& \textbf{\makecell{FLEncoder+\\SL-FLDecoder}} \\
\midrule
\multirow{4}{*}{\centering \textbf{Dataset}}
& \textbf{Channel}
& [1280,16,24,32] & [16,8,8,8] & [256,256,256,256] & [16,16,16,16] & [64,128,256,512] & [32,16,8,8] \\
& \textbf{Depth}
& -- & 16 & -- & 16 & -- & 16 \\
& \textbf{MACs (G)$\downarrow$}
& 0.2311 & \textbf{\underline{0.2248}} & 15.4028 & \textbf{\underline{5.3586}} & 10.3408 & \textbf{\underline{0.7684}} \\
& \textbf{Params (M)$\downarrow$}
& 0.1278 & \textbf{\underline{0.0404}} & 8.5934 & \textbf{\underline{0.1055}} & 4.0604 & \textbf{\underline{0.0506}} \\
\midrule
\multirow{2}{*}{\textbf{Cityscapes}}
& \textbf{RMSE $\downarrow$}
& 0.876 & \textbf{\underline{0.870}} & 0.918 & \textbf{\underline{0.811}} & \textbf{\underline{0.747}} & 0.782 \\
& \textbf{$\delta_1 \uparrow$}
& 0.576 & \textbf{\underline{0.588}} & 0.556 & \textbf{\underline{0.620}} & \textbf{\underline{0.652}} & 0.641 \\
\midrule

\multirow{2}{*}{\textbf{Make3D}}
& \textbf{RMSE $\downarrow$}
& 1.485 & \textbf{\underline{1.434}} & \textbf{\underline{1.457}} & 1.690 & 1.339 & \textbf{\underline{1.324}} \\
& \textbf{$\delta_1 \uparrow$}
& 0.311 & \textbf{\underline{0.330}} & 0.336 & \textbf{\underline{0.342}} & \textbf{\underline{0.369}} & 0.362 \\

\midrule

\multirow{2}{*}{\textbf{DIODE}}
& \textbf{RMSE $\downarrow$}
& 0.572 & \textbf{\underline{0.567}} & 0.555 & \textbf{\underline{0.528}} & 0.595 & \textbf{\underline{0.591}} \\
& \textbf{$\delta_1 \uparrow$}
& 0.217 & \textbf{\underline{0.236}} & 0.271 & \textbf{\underline{0.287}} & 0.236 & \textbf{\underline{0.241}} \\

\midrule

\multirow{2}{*}{\textbf{KITTI}}
& \textbf{RMSE $\downarrow$}
& 2.144 & \textbf{\underline{2.137}} & \textbf{\underline{2.114}} & 2.185 & 2.722 & \textbf{\underline{2.642}} \\
& \textbf{$\delta_1 \uparrow$}
& 0.220 & \textbf{\underline{0.233}} & 0.239 & \textbf{\underline{0.256}} & 0.208 & \textbf{\underline{0.231}} \\

\midrule

\multirow{2}{*}{\textbf{NYU}}
& \textbf{RMSE $\downarrow$}
& \textbf{\underline{1.449}} & 1.559 & 1.501 & \textbf{\underline{1.333}} & 1.862 & \textbf{\underline{1.597}} \\
& \textbf{$\delta_1 \uparrow$}
& 0.450 & \textbf{\underline{0.466}} & 0.444 & \textbf{\underline{0.486}} & 0.353 & \textbf{\underline{0.414}} \\

\midrule

\multirow{2}{*}{\textbf{MODEST}}
& \textbf{RMSE $\downarrow$}
& \textbf{\underline{0.484}} & 0.507 & 0.450 & \textbf{\underline{0.439}} & 0.289 & \textbf{\underline{0.245}} \\
& \textbf{$\delta_1 \uparrow$}
& \textbf{\underline{0.673}} & 0.651 & 0.752 & \textbf{\underline{0.764}} & 0.953 & \textbf{\underline{0.959}} \\

\bottomrule
\end{tabular}
}
\label{Tab:dataset_model_complexity}
\end{table}

\textbf{Spatial Lifting for Depth Estimation.}
We evaluate three encoders (MobileNet~\cite{sandler2018mobilenetv2}, ResNeXt~\cite{xie2017aggregated}, and FLEncoder~\cite{jia2023joint}) combined with three decoders (FastDepth~\cite{icra_2019_fastdepth}, MiDaS~\cite{ranftl2020towards}, and FLDecoder~\cite{jia2023joint}). SL is integrated into all configurations to assess its generality. Quantitative results are reported in Table~\ref{Tab:dataset_model_complexity}. Across all configurations, SL substantially reduces parameters and GMACs in most configurations in both MACs and parameter counts. For instance, under the ResNeXt + MiDaS setting, MACs are reduced by 65.3\% and parameters to 1.2\% of the original model.

% For depth estimation, we evaluate three encoder architectures, MobileNet~\cite{sandler2018mobilenetv2}, ResNeXt~\cite{xie2017aggregated}, and FLEncoder~\cite{jia2023joint}, combined with three decoders, FastDepth~\cite{icra_2019_fastdepth}, MiDaS~\cite{ranftl2020towards}, and FLDecoder~\cite{jia2023joint}. The SL is incorporated into each configuration to assess its generality across architectural designs. Quantitative results are reported in Table~\ref{Tab:depth estimation}. Across all configurations, SL consistently reduces computational cost, leading to substantial decreases in both MACs and parameter counts (see Table~\ref{Tab:arch_details}). For instance, under the ResNeXt + MiDaS setting, MACs are reduced by 65.3\%, while parameters are compressed to 1.2\% of the original model, indicating a significant improvement in efficiency.

Beyond efficiency, a more nuanced performance pattern can be observed. For lightweight architectures such as MobileNet + FastDepth, SL provides stable improvements across nearly all datasets, indicating its robustness in resource-constrained settings. In contrast, for more powerful backbones like ResNeXt + MiDaS, SL not only improves $\delta_1$ accuracy consistently across all datasets but also significantly reduces RMSE on most benchmarks, suggesting that SL enhances both global structural consistency and local depth precision. Although slight performance degradation is observed on specific datasets such as KITTI and Make3D, the overall trend still favors SL-augmented models. For the FSLNet-based architecture, which is already optimized for efficiency, SL exhibits a more selective improvement pattern. While gains in RMSE and $\delta_1$ are evident on datasets such as NYU, KITTI, and MODEST, marginal trade-offs appear on Cityscapes, implying that the benefit of SL may depend on the redundancy level and representational capacity of the baseline architecture. 
Overall, these results indicate that SL provides a consistent accuracy-efficiency trade-off across diverse architectures and datasets. This suggests its applicability as a general module for dense prediction tasks beyond semantic segmentation.

\begin{table}[t]
\footnotesize
\centering
\caption{Correlation between slice-consistency quality score from SL-UNet (8L, 2Res) and ground-truth Dice scores, illustrating the relationship between the prediction quality score $Q$ and the actual Dice performance.}
\resizebox{0.8\linewidth}{!}{
\begin{tabular}{lccccc}
\toprule
\textbf{Dataset} & \textbf{Quality score ($Q$)} & \textbf{Actual Dice} & \textbf{Pearson $r$} & \textbf{Spearman $\rho$} & \textbf{$p$-value} \\
\midrule
CHASE\_DB1   & 78.83 & 72.39 & 0.72 & 0.72 & $> 0.05$ \\
DSB2018      & 91.77 & 88.91 & 0.62 & 0.80 & $< 0.05$ \\
Kvasir-SEG   & 71.55 & 76.03 & 0.71 & 0.71 & $< 0.05$ \\
MoNuSAC      & 76.44 & 70.95 & 0.77 & 0.60 & $< 0.05$ \\
ISIC2018     & 83.48 & 82.97 & 0.73 & 0.67 & $< 0.05$ \\
MoNuSeg      & 80.45 & 79.02 & 0.65 & 0.47 & $< 0.05$ \\
Sartorius    & 72.48 & 73.02 & 0.72 & 0.78 & $< 0.05$ \\
TNBC         & 75.51 & 69.69 & 0.55 & 0.67 & $< 0.05$ \\
FNC          & 75.28 & 68.83 & 0.52 & 0.78 & $< 0.05$ \\
GlaS         & 87.14 & 84.61 & 0.51 & 0.63 & $< 0.05$ \\
BCCD         & 95.95 & 95.70 & 0.87 & 0.75 & $< 0.05$ \\
FIVES        & 79.42 & 79.05 & 0.88 & 0.88 & $< 0.05$ \\
NuInsSeg     & 72.42 & 65.96 & 0.69 & 0.70 & $< 0.05$ \\
\bottomrule
\end{tabular}
}
\label{Tab:correlation}
\end{table}

\begin{figure}[t]
  \centering
  \includegraphics[width=1\textwidth]{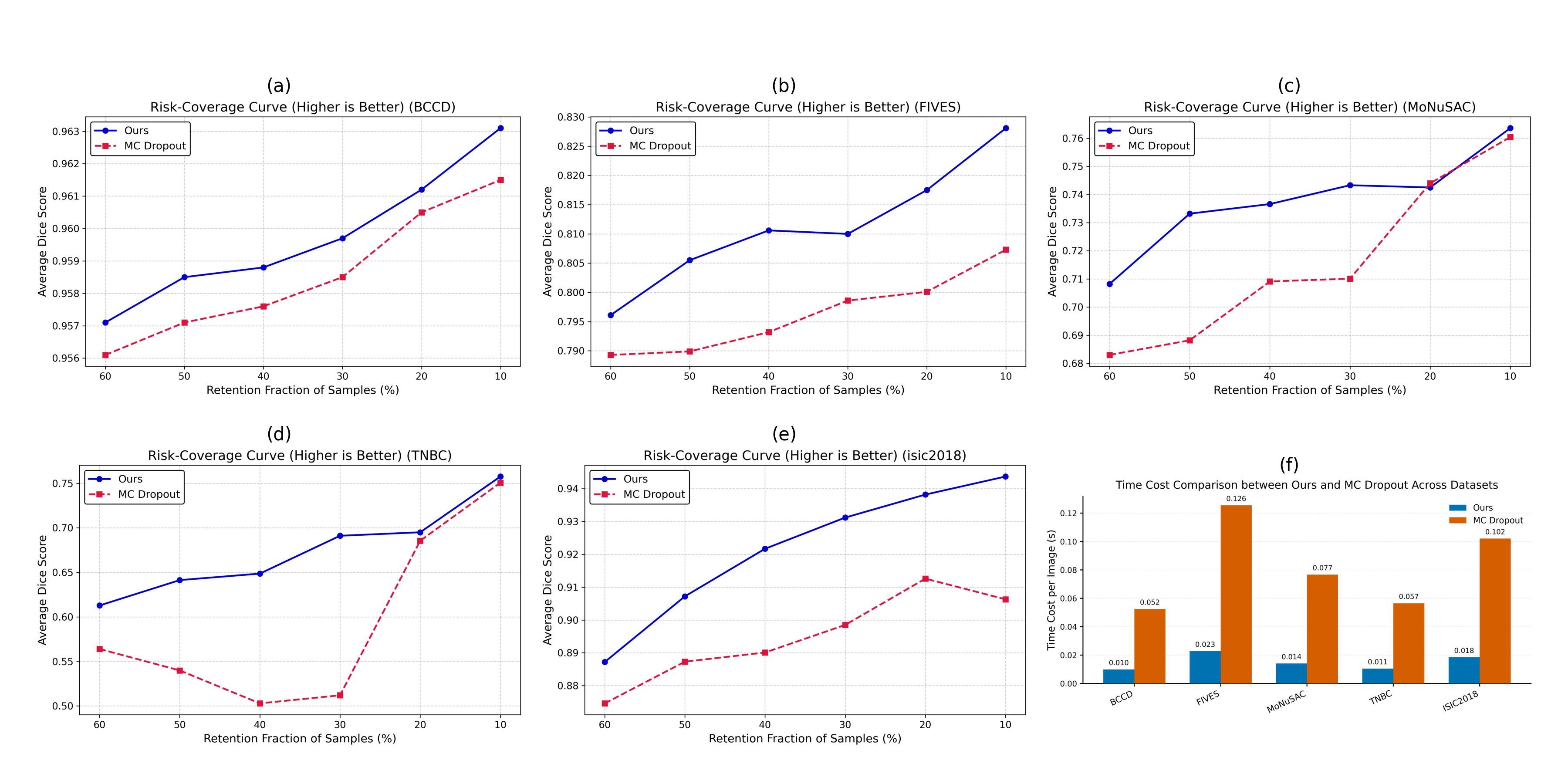}
  \caption{Risk–coverage performance and computational cost comparison between the proposed method and MC Dropout. (a)–(e) Risk–coverage curves (mean Dice vs. retention fraction) on five datasets: BCCD, FIVES, MoNuSAC, TNBC, and ISIC2018. (f) Inference time per image (seconds).} 
  \label{fig:RCC_Time}
\end{figure}

\phantomsection
\label{sec:quality_effectiveness}

\textbf{Validation of $Q$ as a Proxy for Segmentation Quality.}
As shown in Table~\ref{Tab:correlation}, higher values of \(Q\) are generally associated with higher ground-truth Dice,
indicating that slice consistency provides a useful proxy for prediction quality.  Most datasets show statistically significant positive correlations between \(Q\) and Dice, confirming
that slice consistency is informative of prediction quality.

% As shown in Table~\ref{Tab:correlation}, all datasets achieve a Pearson correlation $r > 0.5$, with several cases (e.g., FIVES) reaching values above 0.8. All correlations are statistically significant ($p < 0.05$), indicating that the observed relationship between $Q$ and Dice is unlikely to be due to chance. We further observe that $Q$ is generally consistent with the ground-truth Dice scores, suggesting that it can reasonably reflect segmentation quality in practice. This is particularly evident on datasets such as FIVES, where the correlation reaches 0.88 and the Dice scores (79.42 vs. 79.05), showing strong agreement between $Q$ and actual performance.These results suggest that $Q$ can serve as a practical proxy for segmentation quality in scenarios without full ground-truth annotations, enabling approximate quality assessment in real-world applications.

\textbf{Uncertainty Estimation via Inter-slice Consistency.}
The proposed uncertainty estimation does not rely on Ground Truth (GT) masks, but instead exploits inter-slice prediction consistency as a spatial prior. Inconsistencies across slices are used as an indicator of model uncertainty. In contrast, MC Dropout estimates uncertainty via multiple stochastic forward passes, incurring additional computational cost. We evaluate uncertainty quality using Risk–Coverage Curves (RCC) on five datasets in Fig.~\ref{fig:RCC_Time}(a)–(e), which assess ranking performance by progressively removing samples with higher uncertainty. As shown, our method consistently yields higher Dice at lower coverage, indicating more reliable uncertainty ranking. MC Dropout shows weaker and less stable trends. For a fair comparison, MC Dropout is evaluated with 5 stochastic forward passes, while our method requires only one forward pass (aggregation over the 5 selected slices), resulting in over $5\times$ lower inference time (see Fig.~\ref{fig:RCC_Time}(f)).

\section{Conclusion}
In this work, we introduced Spatial Lifting (SL), a novel modeling paradigm for dense prediction tasks that counterintuitively lifts standard 2D inputs into a higher-dimensional space for processing with 3D architectures. Our evaluations demonstrate that this approach not only achieves highly competitive performance on benchmarks like semantic segmentation and depth estimation while reducing model parameters and computational overhead, but also naturally yields a robust, nearly zero-cost quality estimation metric at inference time by leveraging the variance across prediction slices. SL offers a highly efficient alternative to traditional uncertainty quantification methods like Deep Ensembles or Monte Carlo Dropout. Spatial Lifting offers a scalable pathway toward more reliable deep networks for real-world dense prediction. Future work will further explore lightweight and learnable lifting modules to map inputs to the initial states.

\bibliographystyle{unsrt}
\bibliography{reference}

@inproceedings{touvron2021training,
  title={Training data-efficient image transformers \& distillation through attention},
  author={Touvron, Hugo and Cord, Matthieu and Douze, Matthijs and Massa, Francisco and Sablayrolles, Alexandre and J{\'e}gou, Herv{\'e}},
  booktitle={International conference on machine learning},
  pages={10347--10357},
  year={2021},
  organization={PMLR}
}

@article{krishnamoorthi2018quantizing,
  title={Quantizing deep convolutional networks for efficient inference: A whitepaper},
  author={Krishnamoorthi, Raghuraman},
  journal={arXiv preprint arXiv:1806.08342},
  year={2018}
}

@article{howard2017mobilenets,
  title={Mobilenets: Efficient convolutional neural networks for mobile vision applications},
  author={Howard, Andrew G and Zhu, Menglong and Chen, Bo and Kalenichenko, Dmitry and Wang, Weijun and Weyand, Tobias and Andreetto, Marco and Adam, Hartwig},
  journal={arXiv preprint arXiv:1704.04861},
  year={2017}
}

@inproceedings{zhang2018shufflenet,
  title={Shufflenet: An extremely efficient convolutional neural network for mobile devices},
  author={Zhang, Xiangyu and Zhou, Xinyu and Lin, Meng-Hao and Sun, Jian},
  booktitle={Proceedings of the IEEE conference on computer vision and pattern recognition},
  pages={6848--6856},
  year={2018}
}

@inproceedings{tan2019efficientnet,
  title={Efficientnet: Rethinking model scaling for convolutional neural networks},
  author={Tan, Mingxing and Le, Quoc V},
  booktitle={International conference on machine learning},
  pages={6105--6114},
  year={2019},
  organization={PMLR}
}

@inproceedings{han2015learning,
  title={Learning both weights and connections for efficient neural network},
  author={Han, Song and Pool, Jeff and Tran, John and Dally, William},
  booktitle={Advances in neural information processing systems},
  volume = {28},
  year={2015}
}

@article{frankle2018lottery,
  title={The lottery ticket hypothesis: Finding sparse, trainable neural networks},
  author={Frankle, Jonathan and Carbin, Michael},
  journal={arXiv preprint arXiv:1803.03635},
  year={2018}
}

@article{hinton2015distilling,
  title={Distilling the knowledge in a neural network},
  author={Hinton, Geoffrey and Vinyals, Oriol and Dean, Jeff},
  journal={arXiv preprint arXiv:1503.02531},
  year={2015}
}

@article{liu2018darts,
  title={Darts: Differentiable architecture search},
  author={Liu, Hanxiao and Simonyan, Karen and Yang, Yiming},
  journal={arXiv preprint arXiv:1806.09055},
  year={2018}
}

@inproceedings{tan2019mnasnet,
  title={Mnasnet: Platform-aware neural architecture search for mobile},
  author={Tan, Mingxing and Chen, Bo and Pang, Ruoming and Vasudevan, Vijay and Sandler, Mark and Howard, Andrew and Le, Quoc V},
  booktitle={Proceedings of the IEEE/CVF conference on computer vision and pattern recognition},
  pages={2820--2828},
  year={2019}
}

@article{cai2018proxylessnas,
  title={Proxylessnas: Direct neural architecture search on target task and hardware},
  author={Cai, Han and Zhu, Ligeng and Han, Song},
  journal={arXiv preprint arXiv:1812.00332},
  year={2018}
}

@inproceedings{Ronneberger2015_UNet,
  title={U-net: Convolutional networks for biomedical image segmentation},
  author={Ronneberger, Olaf and Fischer, Philipp and Brox, Thomas},
  booktitle={Medical image computing and computer-assisted intervention--MICCAI 2015: 18th international conference, Munich, Germany, October 5-9, 2015, proceedings, part III 18},
  pages={234--241},
  year={2015},
  organization={Springer}
}

@article{Chen2017_DeepLab,
  title={Rethinking atrous convolution for semantic image segmentation},
  author={Chen, Liang-Chieh and Papandreou, George and Schroff, Florian and Adam, Hartwig},
  journal={arXiv preprint arXiv:1706.05587},
  year={2017}
}

@inproceedings{Bhat2021_AdaBins,
  title={Adabins: Depth estimation using adaptive bins},
  author={Bhat, Shariq Farooq and Alhashim, Ibraheem and Wonka, Peter},
  booktitle={Proceedings of the IEEE/CVF conference on computer vision and pattern recognition},
  pages={4009--4018},
  year={2021}
}

@inproceedings{Long2015_FCN,
  title={Fully convolutional networks for semantic segmentation},
  author={Long, Jonathan and Shelhamer, Evan and Darrell, Trevor},
  booktitle={Proceedings of the IEEE conference on computer vision and pattern recognition},
  pages={3431--3440},
  year={2015}
}

@article{Dosovitskiy2021_ViT,
  title={An image is worth 16x16 words: Transformers for image recognition at scale},
  author={Dosovitskiy, Alexey and Beyer, Lucas and Kolesnikov, Alexander and Weissenborn, Dirk and Zhai, Xiaohua and Unterthiner, Thomas and Dehghani, Mostafa and Minderer, Matthias and Heigold, Georg and Gelly, Sylvain and others},
  journal={arXiv preprint arXiv:2010.11929},
  year={2020}
}

@inproceedings{Gal2016_DropoutUncertainty,
  title={Dropout as a bayesian approximation: Representing model uncertainty in deep learning},
  author={Gal, Yarin and Ghahramani, Zoubin},
  booktitle={International Conference on Machine Learning},
  pages={1050--1059},
  year={2016},
  organization={PMLR}
}

@article{wang2022pvt,
  title={Pvt v2: Improved baselines with pyramid vision transformer},
  author={Wang, Wenhai and Xie, Enze and Li, Xiang and Fan, Deng-Ping and Song, Kaitao and Liang, Ding and Lu, Tong and Luo, Ping and Shao, Ling},
  journal={Computational Visual Media},
  volume={8},
  number={3},
  pages={415--424},
  year={2022},
  publisher={Springer}
}

@inproceedings{Cicek2016_3DUNet,
  title={3D U-Net: learning dense volumetric segmentation from sparse annotation},
  author={{\c{C}}i{\c{c}}ek, {\"O}zg{\"u}n and Abdulkadir, Ahmed and Lienkamp, Soeren S and Brox, Thomas and Ronneberger, Olaf},
  booktitle={Medical Image Computing and Computer-Assisted Intervention--MICCAI 2016: 19th International Conference, Athens, Greece, October 17-21, 2016, Proceedings, Part II 19},
  pages={424--432},
  year={2016},
  organization={Springer}
}

@article{He2015_SPPNet,
  title={Spatial pyramid pooling in deep convolutional networks for visual recognition},
  author={He, Kaiming and Zhang, Xiangyu and Ren, Shaoqing and Sun, Jian},
  journal={IEEE Transactions on Pattern Analysis and Machine Intelligence},
  volume={37},
  number={9},
  pages={1904--1916},
  year={2015},
  publisher={IEEE}
}

@InProceedings{Zhao2017_PSPNet,
    author = {Zhao, Hengshuang and Shi, Jianping and Qi, Xiaojuan and Wang, Xiaogang and Jia, Jiaya},
    title = {Pyramid Scene Parsing Network},
    booktitle = {Proceedings of the IEEE Conference on Computer Vision and Pattern Recognition (CVPR)},
    month = {July},
    year = {2017}
}

@InProceedings{Zheng2021_SETR,
    author    = {Zheng, Sixiao and Lu, Jiachen and Zhao, Hengshuang and Zhu, Xiatian and Luo, Zekun and Wang, Yabiao and Fu, Yanwei and Feng, Jianfeng and Xiang, Tao and Torr, Philip H.S. and Zhang, Li},
    title     = {Rethinking Semantic Segmentation From a Sequence-to-Sequence Perspective With Transformers},
    booktitle = {Proceedings of the IEEE/CVF Conference on Computer Vision and Pattern Recognition (CVPR)},
    month     = {June},
    year      = {2021},
    pages     = {6881-6890}
}

@article{shi2016real,
  title={Real-Time Single Image and Video Super-Resolution Using an Efficient Sub-Pixel Convolutional Neural Network},
  author={Wenzhe Shi and Jose Caballero and Ferenc Husz{\'a}r and Johannes Totz and Andrew P. Aitken and Rob Bishop and Daniel Rueckert and Zehan Wang},
  journal={2016 IEEE Conference on Computer Vision and Pattern Recognition (CVPR)},
  year={2016},
  pages={1874-1883},
}

@book{de2000computational,
  title={Computational geometry: algorithms and applications},
  author={De Berg, Mark},
  year={2000},
  publisher={Springer Science \& Business Media}
}

@inproceedings{He2016_ResNet,
  title={Deep residual learning for image recognition},
  author={He, Kaiming and Zhang, Xiangyu and Ren, Shaoqing and Sun, Jian},
  booktitle={Proceedings of the IEEE Conference on Computer Vision and Pattern Recognition},
  pages={770--778},
  year={2016}
}

@inproceedings{Xie2021_SegFormer,
  title={SegFormer: Simple and efficient design for semantic segmentation with transformers},
  author={Xie, Enze and Wang, Wenhai and Yu, Zhiding and Anandkumar, Anima and Alvarez, Jose M and Luo, Ping},
  booktitle={Advances in Neural Information Processing Systems},
  volume={34},
  pages={12077--12090},
  year={2021}
}

@article{fraz2012ensemble,
  title={An ensemble classification-based approach applied to retinal blood vessel segmentation},
  author={Fraz, Muhammad Moazam and Remagnino, Paolo and Hoppe, Andreas and Uyyanonvara, Bunyarit and Rudnicka, Alicja R and Owen, Christopher G and Barman, Sarah A},
  journal={IEEE Transactions on Biomedical Engineering},
  volume={59},
  number={9},
  pages={2538--2548},
  year={2012},
  publisher={IEEE}
}

@article{Caicedo2019NucleusSA,
  title={Nucleus segmentation across imaging experiments: the 2018 Data Science Bowl},
  author={Juan C. Caicedo and Allen Goodman and Kyle W. Karhohs and Beth A. Cimini and Jeanelle Ackerman and Marzieh Haghighi and Cherkeng Heng and Tim Becker and Minh Doan and Claire McQuin and Mohammad Hossein Rohban and Shantanu Singh and Anne E Carpenter},
  journal={Nature Methods},
  year={2019},
  volume={16},
  pages={1247 - 1253},
}

@inproceedings{jha2020kvasir,
  title={Kvasir-seg: A segmented polyp dataset},
  author={Jha, Debesh and Smedsrud, Pia H and Riegler, Michael A and Halvorsen, P{\aa}l and De Lange, Thomas and Johansen, Dag and Johansen, H{\aa}vard D},
  booktitle={International conference on multimedia modeling},
  pages={451--462},
  year={2019},
  organization={Springer}
}

@article{verma2021monusac2020,
  title={MoNuSAC2020: A multi-organ nuclei segmentation and classification challenge},
  author={Verma, Ruchika and Kumar, Neeraj and Patil, Abhijeet and Kurian, Nikhil Cherian and Rane, Swapnil and Graham, Simon and Vu, Quoc Dang and Zwager, Mieke and Raza, Shan E Ahmed and Rajpoot, Nasir and others},
  journal={IEEE Transactions on Medical Imaging},
  volume={40},
  number={12},
  pages={3413--3423},
  year={2021},
  publisher={IEEE}
}

@article{kumar2017dataset,
  title={A dataset and a technique for generalized nuclear segmentation for computational pathology},
  author={Kumar, Neeraj and Verma, Ruchika and Sharma, Sanuj and Bhargava, Surabhi and Vahadane, Abhishek and Sethi, Amit},
  journal={IEEE Transactions on Medical Imaging},
  volume={36},
  number={7},
  pages={1550--1560},
  year={2017},
  publisher={IEEE}
}

@misc{sartorius-cell-instance-segmentation,
    author = {Addison Howard and Ashley Chow and CorporateResearchSartorius and Maria Ca and Phil Culliton and Tim Jackson},
    title = {Sartorius - Cell Instance Segmentation},
    year = {2021},
    howpublished = {\url{https://kaggle.com/competitions/sartorius-cell-instance-segmentation}},
    note = {Accessed: 2025-05-24}
}

@article{8438559,
  title={Segmentation of nuclei in histopathology images by deep regression of the distance map},
  author={Naylor, Peter and La{\'e}, Marick and Reyal, Fabien and Walter, Thomas},
  journal={IEEE Transactions on Medical Imaging},
  volume={38},
  number={2},
  pages={448--459},
  year={2018},
  publisher={IEEE}
}

@article{mahbod2023nuinsseg,
  title={Nuinsseg: a fully annotated dataset for nuclei instance segmentation in h\&e-stained histological images},
  author={Mahbod, Amirreza and Polak, Christine and Feldmann, Katharina and Khan, Rumsha and Gelles, Katharina and Dorffner, Georg and Woitek, Ramona and Hatamikia, Sepideh and Ellinger, Isabella},
  journal={Scientific Data},
  volume={11},
  number={1},
  pages={295},
  year={2024},
  publisher={Nature Publishing Group UK London}
}

@article{jin2022fives,
  title={Fives: A fundus image dataset for artificial Intelligence based vessel segmentation},
  author={Jin, Kai and Huang, Xingru and Zhou, Jingxing and Li, Yunxiang and Yan, Yan and Sun, Yibao and Zhang, Qianni and Wang, Yaqi and Ye, Juan},
  journal={Scientific Data},
  volume={9},
  number={1},
  pages={475},
  year={2022},
  publisher={Nature Publishing Group UK London}
}

@inproceedings{codella2018skin,
  title={Skin lesion analysis toward melanoma detection: A challenge at the 2017 international symposium on biomedical imaging (isbi), hosted by the international skin imaging collaboration (isic)},
  author={Codella, Noel CF and Gutman, David and Celebi, M Emre and Helba, Brian and Marchetti, Michael A and Dusza, Stephen W and Kalloo, Aadi and Liopyris, Konstantinos and Mishra, Nabin and Kittler, Harald and others},
  booktitle={2018 IEEE 15th International Symposium on Biomedical Imaging (ISBI 2018)},
  pages={168--172},
  year={2018},
  organization={IEEE}
}

@article{tschandl2018ham10000,
  title={The HAM10000 dataset, a large collection of multi-source dermatoscopic images of common pigmented skin lesions},
  author={Tschandl, Philipp and Rosendahl, Cliff and Kittler, Harald},
  journal={Scientific Data},
  volume={5},
  number={1},
  pages={1--9},
  year={2018},
  publisher={Nature Publishing Group}
}

@misc{Sarker2023,
    title={Blood Cell Segmentation Dataset},
    DOI={10.34740/KAGGLE/DSV/6107556},
    publisher={Kaggle},
    author={Deponker Sarker Depto and Shazidur Rahman and Md. Mekayel Hosen and Mst Shapna Akter and Tamanna Rahman Reme and Aimon Rahman and Hasib Zunai and M.R.C.Mahdy and M. Sohel Rahman and Jeet B Lahiri},
    year={2023},
    howpublished = {\url{https://www.kaggle.com/dsv/6107556}},
    note = {Accessed: 2025-05-24}
}

@article{morelli2021automating,
  title={Automating cell counting in fluorescent microscopy through deep learning with c-ResUnet},
  author={Morelli, Roberto and Clissa, Luca and Amici, Roberto and Cerri, Matteo and Hitrec, Timna and Luppi, Marco and Rinaldi, Lorenzo and Squarcio, Fabio and Zoccoli, Antonio},
  journal={Scientific Reports},
  volume={11},
  number={1},
  pages={22920},
  year={2021},
  publisher={Nature Publishing Group UK London}
}

@article{hitrec2019neural,
  title={Neural control of fasting-induced torpor in mice},
  author={Hitrec, Timna and Luppi, Marco and Bastianini, Stefano and Squarcio, Fabio and Berteotti, Chiara and Lo Martire, Viviana and Martelli, Davide and Occhinegro, Alessandra and Tupone, Domenico and Zoccoli, Giovanna and others},
  journal={Scientific Reports},
  volume={9},
  number={1},
  pages={15462},
  year={2019},
  publisher={Nature Publishing Group UK London}
}

@article{sirinukunwattana2017gland,
  title={Gland segmentation in colon histology images: The glas challenge contest},
  author={Sirinukunwattana, Korsuk and Pluim, Josien PW and Chen, Hao and Qi, Xiaojuan and Heng, Pheng-Ann and Guo, Yun Bo and Wang, Li Yang and Matuszewski, Bogdan J and Bruni, Elia and Sanchez, Urko and others},
  journal={Medical Image Analysis},
  volume={35},
  pages={489--502},
  year={2017},
  publisher={Elsevier}
}

@article{sirinukunwattana2015stochastic,
  title={A stochastic polygons model for glandular structures in colon histology images},
  author={Sirinukunwattana, Korsuk and Snead, David RJ and Rajpoot, Nasir M},
  journal={IEEE Transactions on Medical Imaging},
  volume={34},
  number={11},
  pages={2366--2378},
  year={2015},
  publisher={IEEE}
}

@article{ZHANG2022106173,
    title = {HSNet: A hybrid semantic network for polyp segmentation},
    journal = {Computers in Biology and Medicine},
    volume = {150},
    pages = {106173},
    year = {2022},
    issn = {0010-4825},
    author = {Wenchao Zhang and Chong Fu and Yu Zheng and Fangyuan Zhang and Yanli Zhao and Chiu-Wing Sham},
}

@InProceedings{rahman2024emcad,
    author    = {Rahman, Md Mostafijur and Munir, Mustafa and Marculescu, Radu},
    title     = {EMCAD: Efficient Multi-scale Convolutional Attention Decoding for Medical Image Segmentation},
    booktitle = {Proceedings of the IEEE/CVF Conference on Computer Vision and Pattern Recognition (CVPR)},
    month     = {June},
    year      = {2024},
    pages     = {11769-11779}
}

@inproceedings{cordts2016cityscapes,
  title={The cityscapes dataset for semantic urban scene understanding},
  author={Cordts, Marius and Omran, Mohamed and Ramos, Sebastian and Rehfeld, Timo and Enzweiler, Markus and Benenson, Rodrigo and Franke, Uwe and Roth, Stefan and Schiele, Bernt},
  booktitle={Proceedings of the IEEE conference on computer vision and pattern recognition},
  pages={3213--3223},
  year={2016}
}

@inproceedings{liu2019end,
  title={End-to-end multi-task learning with attention},
  author={Liu, Shikun and Johns, Edward and Davison, Andrew J},
  booktitle={Proceedings of the IEEE/CVF conference on computer vision and pattern recognition},
  pages={1871--1880},
  year={2019}
}

@article{saxena2005learning,
  title={Learning depth from single monocular images},
  author={Saxena, Ashutosh and Chung, Sung and Ng, Andrew},
  journal={Advances in neural information processing systems},
  volume={18},
  year={2005}
}

@inproceedings{saxena2007learning,
  title={Learning 3-d scene structure from a single still image},
  author={Saxena, Ashutosh and Sun, Min and Ng, Andrew Y},
  booktitle={2007 IEEE 11th international conference on computer vision},
  pages={1--8},
  year={2007},
  organization={IEEE}
}

@article{vasiljevic2019diode,
  title={DIODE: A Dense Indoor and Outdoor DEpth Dataset},
  author={Igor Vasiljevic and Nicholas I. Kolkin and Shanyi Zhang and Ruotian Luo and Haochen Wang and Falcon Z. Dai and Andrea F. Daniele and Mohammadreza Mostajabi and Steven Basart and Matthew R. Walter and Gregory Shakhnarovich},
  journal={ArXiv},
  year={2019},
  volume={abs/1908.00463},
}

@article{geiger2013vision,
  title={Vision meets robotics: The kitti dataset},
  author={Geiger, Andreas and Lenz, Philip and Stiller, Christoph and Urtasun, Raquel},
  journal={The international journal of robotics research},
  volume={32},
  number={11},
  pages={1231--1237},
  year={2013},
  publisher={Sage Publications Sage UK: London, England}
}

@misc{shan2021modestkaggle,
  author       = {Aman Shan},
  title        = {MODEST Museum Dataset},
  year         = {2021},
  howpublished = {\url{https://www.kaggle.com/datasets/shanwizard/modest-museum-dataset}},
  note         = {Accessed: 2025-05-25}
}

@inproceedings{silberman2012indoor,
  title={Indoor segmentation and support inference from rgbd images},
  author={Silberman, Nathan and Hoiem, Derek and Kohli, Pushmeet and Fergus, Rob},
  booktitle={Computer Vision--ECCV 2012: 12th European Conference on Computer Vision, Florence, Italy, October 7-13, 2012, Proceedings, Part V 12},
  pages={746--760},
  year={2012},
  organization={Springer}
}

@inproceedings{sandler2018mobilenetv2,
  title={Mobilenetv2: Inverted residuals and linear bottlenecks},
  author={Sandler, Mark and Howard, Andrew and Zhu, Menglong and Zhmoginov, Andrey and Chen, Liang-Chieh},
  booktitle={Proceedings of the IEEE conference on computer vision and pattern recognition},
  pages={4510--4520},
  year={2018}
}

@article{ranftl2020towards,
  title={Towards robust monocular depth estimation: Mixing datasets for zero-shot cross-dataset transfer},
  author={Ranftl, Ren{\'e} and Lasinger, Katrin and Hafner, David and Schindler, Konrad and Koltun, Vladlen},
  journal={IEEE Transactions on Pattern Analysis and Machine Intelligence},
  volume={44},
  number={3},
  pages={1623--1637},
  year={2020},
  publisher={IEEE}
}

@inproceedings{xie2017aggregated,
  title={Aggregated residual transformations for deep neural networks},
  author={Xie, Saining and Girshick, Ross and Doll{\'a}r, Piotr and Tu, Zhuowen and He, Kaiming},
  booktitle={Proceedings of the IEEE Conference on Computer Vision and Pattern Recognition (CVPR)},
  pages={1492--1500},
  year={2017}
}

@inproceedings{icra_2019_fastdepth,
  title={Fastdepth: Fast monocular depth estimation on embedded systems},
  author={Wofk, Diana and Ma, Fangchang and Yang, Tien-Ju and Karaman, Sertac and Sze, Vivienne},
  booktitle={2019 International Conference on Robotics and Automation (ICRA)},
  pages={6101--6108},
  year={2019},
  organization={IEEE}
}

@article{jia2023joint,
  title={Joint learning of frequency and spatial domains for dense image prediction},
  author={Jia, Shaocheng and Yao, Wei},
  journal={ISPRS Journal of Photogrammetry and Remote Sensing},
  volume={195},
  pages={14--28},
  year={2023},
  publisher={Elsevier}
}

@article{lakshminarayanan2017simple,
  title={Simple and scalable predictive uncertainty estimation using deep ensembles},
  author={Lakshminarayanan, Balaji and Pritzel, Alexander and Blundell, Charles},
  journal={Advances in neural information processing systems},
  volume={30},
  year={2017}
}

@inproceedings{gal2016dropout,
  title={Dropout as a bayesian approximation: Representing model uncertainty in deep learning},
  author={Gal, Yarin and Ghahramani, Zoubin},
  booktitle={international conference on machine learning},
  pages={1050--1059},
  year={2016},
  organization={PMLR}
}

@inproceedings{blundell2015weight,
  title={Weight uncertainty in neural network},
  author={Blundell, Charles and Cornebise, Julien and Kavukcuoglu, Koray and Wierstra, Daan},
  booktitle={International conference on machine learning},
  pages={1613--1622},
  year={2015},
  organization={PMLR}
}

@inproceedings{shanmugam2021better,
  title={Better aggregation in test-time augmentation},
  author={Shanmugam, Divya and Blalock, Davis and Balakrishnan, Guha and Guttag, John},
  booktitle={Proceedings of the IEEE/CVF international conference on computer vision},
  pages={1214--1223},
  year={2021}
}

@article{kendall2017uncertainties,
  title={What uncertainties do we need in bayesian deep learning for computer vision?},
  author={Kendall, Alex and Gal, Yarin},
  journal={Advances in neural information processing systems},
  volume={30},
  year={2017}
}

% \newpage
% \input{checklist.tex}

\newpage
\appendix
\section{Dataset Details}

\noindent\textbf{For Semantic Segmentation,} we conduct experiments on 13 publicly available datasets. These datasets encompass diverse anatomical structures, including the retina, skin, gastrointestinal tract, glands and various cell types, as well as multiple imaging modalities such as fundus photography, dermoscopy, histopathology, endoscopy, and microscopy. Specifically, the datasets are as follows: CHASE\_DB1, DSB2018, Kvasir, MoNuSAC, ISIC2018,MoNuSeg, Sartorius, TNBC, Fluorescent Neuronal Cells, GlaS, BCCD, FIVES, NuInsSeg. The CHASE\_DB1 and FIVES datasets are both used for retinal vessel segmentation. FIVES contains 800 multi-disease fundus photographs annotated at the pixel level, with 600 used for training and 200 for testing. For CHASE\_DB1, we select 21 images for training and 7 images for testing from a set of 28 images. ISIC2018 is a large-scale dataset for skin lesion segmentation, including 2,594 paired training images and 1,000 paired test images. Kvasir\text{-}SEG is an endoscopic dataset designed for pixel-level segmentation of colonic polyps. The dataset consists of 1,000 gastrointestinal polyp images and their corresponding segmentation masks, all annotated and verified by experienced gastroenterologists. We follow the official data split to divide the dataset into 880 images for training and 120 images for validation. GlaS is a gland segmentation dataset, it consists of 85 training images and 80 testing images, each focusing on segmenting glands in colorectal regions. The remaining datasets, including BCCD, DSB2018, Fluorescent Neuronal Cells, MoNuSAC, MoNuSeg, NuInsSeg, Sartorius, and TNBC, are specifically designed for cell or nucleus segmentation tasks. 

\noindent\textbf{For Depth Estimation,} we employ 6 publicly available datasets: Cityscapes, Make3D, DIODE, KITTI, NYU Depth and MODEST. These datasets cover a wide range of scenes, from outdoor urban environments to indoor scenarios, providing diverse depth cues and challenging variations in lighting, texture, and geometry. Such diversity ensures comprehensive evaluation of model performance under various real-world conditions. The outdoor urban scene datasets include Cityscapes, KITTI, DIODE and Make3D. Cityscapes contains high-resolution images captured in urban street scenes, providing stereo image pairs with dense depth annotations. we use 2,975 images for training and 500 images for testing from the official Cityscapes split. KITTI, collected from a moving vehicle in urban and highway settings, offers stereo RGB images with LiDAR-based ground truth depth, we utilize a subset of 1,000 images originally designated for validation, which we further split into 800 images for training and 200 images for testing to evaluate our model. Make3D provides single outdoor images and corresponding laser-scanned depth maps, containing 425 images divided into 375 training and 50 testing samples. The DIODE dataset comprises RGB-D images captured by a handheld sensor in both indoor and outdoor environments. In this work, only the outdoor subset is utilized, with 80\% of the data used for training and 20\% for testing. The indoor scene datasets include NYU Depth V2 and MODEST. NYU Depth V2 is captured with a Microsoft Kinect sensor and consists of RGB-D images from various indoor scenes, with 50688 training images and 654 test images according to the official partition. MODEST is a large-scale synthetic dataset containing approximately 400,000 images, generated by compositing foreground objects onto diverse background scenes in museum environments. It is designed specifically for monocular depth estimation and segmentation, offering complex indoor layouts with high-quality depth annotations. {More details on the datasets used for the semantic segmentation and depth estimation experiments can be found in Table~\ref{Tab:Datasets}.}

\section{Implementation Details}

\textbf{For Semantic Segmentation.} All models were trained for 40 epochs using the AdamW optimizer with a learning rate of 0.001 and a Cosine Annealing Warm Restarts (CAWR) learning rate scheduler. A gradient clipping value of 0.5 was applied, and the loss functions included BCE with IoU for most configurations, while EMCAD and SL-EMCAD utilized weighted BCE (wbce) combined with weighted IoU (wiou). Multi-scale training was applied across all models using scale factors of 0.75, 1.0, and 1.25.(see Table~\ref{Tab:train_details}). 

\textbf{For Depth Estimation.} All models underwent training for 50 epochs using the AdamW optimization method with an initial learning rate of 0.001. To promote stable convergence during training, a stepwise learning rate decay schedule was utilized, and gradient clipping was applied with a maximum norm of 0.5. Distinct from semantic segmentation, the depth estimation task utilized loss functions tailored to the characteristics of sparse ground truth, specifically Masked L1 Loss and Masked Mean Squared Error (MSE) Loss, to effectively exclude invalid or missing depth regions during optimization.(see Table~\ref{Tab:train_details})

\begin{table}[t]
\footnotesize
\centering
\caption{Details of datasets used for experiments: name, task type, modality, training/testing splits, and image resolution.}
\resizebox{\textwidth}{!}{
\begin{tabular}{@{}l l l c c l @{}}
\toprule
\textbf{Name} & \textbf{Task Type} & \textbf{Modality} & \textbf{Training} & \textbf{Testing} & \textbf{Resolution (px)} \\
\midrule
CHASE\_DB1                  & Retinal Vessel Segmentation             & Fundus Photography                       & 21     & 7    & 999$\times$960 \\
FIVES                       & Retinal Vessel Segmentation             & Fundus Photography                       & 600    & 200  & 2048$\times$2048 \\
ISIC2018                     & Skin Lesion Segmentation                & Dermoscopic Imaging                       & 2594   & 1000 & Mixed \\
Kvasir-SEG                   & Colonic Polyp Segmentation              & Endoscopic Imaging                        & 880    & 120  & Mixed \\
GlaS                         & Gland Segmentation                      & H\&E Histopathological Imaging           & 85     & 80   & 775$\times$522 \\
BCCD                         & Cell Segmentation                        & Microscopy Imaging                        & 1169   & 159  & 1600$\times$1200 \\
DSB2018                      & Nucleus Segmentation                     & Microscopy Imaging                        & 600    & 70   & Mixed \\
Fluorescent Neuronal Cells   & Neuronal Cell Segmentation               & Microscopy Imaging                        & 200    & 83   & 1600$\times$1200 \\
MoNuSAC                      & Nucleus Segmentation and Classification  & H\&E Histopathological Imaging           & 209    & 85   & Mixed \\
MoNuSeg                      & Nucleus Segmentation                     & H\&E Histopathological Imaging           & 37     & 14   & 1000$\times$1000 \\
NuInsSeg                     & Nucleus Segmentation                     & H\&E Histopathological Imaging           & 510    & 155  & 512$\times$512 \\
Sartorius                    & Cell Segmentation                        & H\&E Histopathological Imaging           & 540    & 66   & 704$\times$520 \\
TNBC                         & Nucleus Segmentation                     & H\&E Histopathological Imaging           & 36     & 14   & 512$\times$512 \\
Cityscapes                   & Depth Estimation                         & Urban Street Scenes                        & 2975   & 500  & 256$\times$128 \\
Make3D                       & Depth Estimation                         & Outdoor Landscape                          & 375    & 50   & 1704$\times$2272 \\
DIODE                        & Depth Estimation                         & Diverse Environments                        & 386    & 60   & Mixed \\
KITTI                        & Depth Estimation                         & Autonomous Driving Scenes                  & 800    & 200  & 1216$\times$352 \\
MODEST                       & Depth Estimation                         & Driving Scenes                              & 395k   & 5k   & 224$\times$224 \\
NYU Depth                     & Depth Estimation                         & Indoor Scenes                               & 50688  & 654  & 640$\times$480 \\
\bottomrule
\end{tabular}
}
\label{Tab:Datasets}
\end{table}

\begin{table}[t]
\footnotesize
\centering
\caption{Training configurations: semantic segmentation (cosine-annealed learning rate, weighted BCE+IoU losses) and depth estimation (multi-step decay, masked MSE+L1 losses). Common: AdamW optimizer.}
\label{Tab:train_details}
\begin{tabular}{@{}lcc@{}}
\toprule
\textbf{Configuration} & \textbf{Segmentation Models} & \textbf{Depth Estimation Models} \\
\midrule
Input Size     & 512*512 & 224*224  \\
Batch Size     & 1 & 4  \\
Epoch          & 40  & 50 \\
Optimizer      & AdamW & AdamW \\
Learning Rate  & 0.001 & 0.001 \\
Scheduler      & Cosine Annealing Warm Restarts & Stepwise Learning Rate decay \\
Clip           & 0.5 & 0.5 \\
Loss Function  & WBCE + WIoU & MaskedMSE + MaskedL1 \\
\bottomrule
\end{tabular}
\end{table}

\section{Depth-wise Interpretability and Best Slice Selection}
\begin{figure}[t]
  \centering
  \includegraphics[width=0.7\linewidth]{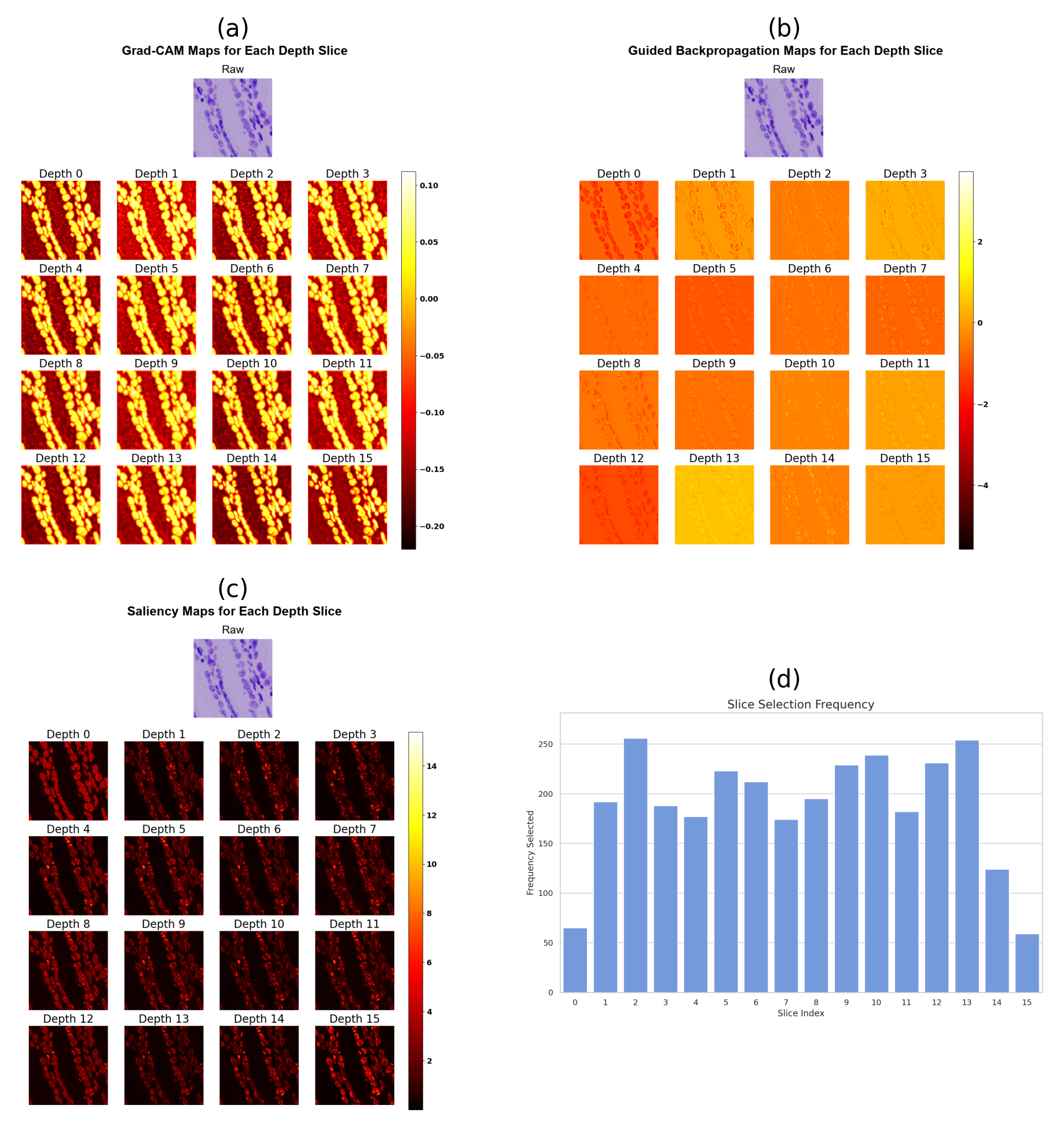}
  \caption{Visualization results using different interpretability methods for each depth slice.}
  \label{fig:visualizations}
\end{figure}

To understand how the model leverages depth-wise information and to demystify the slice selection mechanism, we utilize three distinct interpretability techniques alongside a quantitative frequency analysis, as presented in Fig.~\ref{fig:visualizations}. For illustration, we take the DSB2018 dataset with $m = 16$ as a representative example. Specifically, \textbf{Fig.~\ref{fig:visualizations}(a) (Grad-CAM)} highlights coarse, region-level areas of high activation, indicating where the model is broadly focusing its attention. \textbf{Fig.~\ref{fig:visualizations}(b) (Guided Backpropagation)} captures fine-grained, high-resolution details, emphasizing the specific structural edges that positively contribute to the prediction. \textbf{Fig.~\ref{fig:visualizations}(c) (Saliency Maps)} reveals pixel-level gradient sensitivities, highlighting the fundamental spatial features deemed most critical by the network. Finally, \textbf{Fig.~\ref{fig:visualizations}(d)} provides a statistical histogram showing the frequency with which each depth slice (from index 0 to 15) is automatically selected by the model as the most informative representation.

\textbf{Visualization Analysis Guided by Selection Frequency.} 
A joint analysis of these figures reveals a clear operational pattern. The histogram in Fig.~\ref{fig:visualizations}(d) demonstrates a distinct bell-shaped distribution: the intermediate depth slices (e.g., slices 2 through 13) are selected with high frequency, whereas the extreme boundary slices (Depth 0 and Depth 15) are rarely chosen. 
When correlating this quantitative trend with the qualitative interpretability maps, we observe that while the model maintains a relatively stable global perception across all depths (as evidenced by the structural consistency in the Grad-CAM maps), the quality of fine-grained feature extraction varies. In both the Saliency Maps and Guided Backpropagation maps, the boundary slices (e.g., Depth 0) exhibit somewhat weaker activation intensities and less defined structural outlines. In contrast, the frequently selected intermediate slices show much stronger, sharper, and more precisely concentrated gradient responses on critical tissue edges, explaining why the network naturally favors them.

\textbf{Mechanistic Explanation: Boundary Padding Effects in Spatial Lifting.} 
The network's heavy preference for intermediate slices and avoidance of boundary slices is a direct consequence of the \textit{Slifting} architectural design. During the depth-wise interaction and feature extraction phases within the Slifting blocks, convolutional operations require padding at the depth boundaries ($z=0$ and $z=15$). The standard practice of zero-padding at these extreme ends restricts the effective receptive field in the depth dimension and introduces artificial boundary artifacts. Consequently, the boundary slices lack complete, bi-directional contextual information. Central slices, on the other hand, do not suffer from this boundary truncation; they benefit from symmetric spatial contexts from adjacent layers above and below them, allowing the Slifting operations to extract far more coherent, robust, and semantically rich features.

\textbf{Automatic Top-$s$ Slice Selection.} 
Guided by these intrinsic dynamics, our model's strategy to automatically select and fuse the predictions from the Top-$s$ scoring slices is highly rational. By systematically discarding the contextually degraded boundary slices (which suffer from zero-padding artifacts) and dynamically focusing on the highly informative representations from the central depths, the model inherently filters out boundary noise. This mechanism ensures that the final fused prediction relies strictly on the most stable and feature-rich cross-sections, thereby achieving superior performance.

\section{Impact of Network Hyperparameters \texorpdfstring{$(m, C_{*}, s)$}{(m, C*, s)} on MACs, Parameter Count, and Performance}
In this section, we conduct an extensive ablation study to evaluate how different hyperparameter configurations $(m, C_*, s)$ affect model complexity, computational efficiency, and segmentation performance across 13 datasets. The baseline for comparison is a standard UNet with 8 layers and 2 residual blocks per layer.

We first compare the lightweight configuration $(16, 4, 5)$ with the baseline UNet (8L, 2Res). The baseline model contains 43.2355M parameters and requires 20.7387G MACs, whereas $(16, 4, 5)$ reduces the parameter count by 99.97\% and the computational cost by 87.96\%. However, such extreme compression leads to noticeable degradation in segmentation performance, indicating that overly compact representations are insufficient for accurate prediction. To address this issue, we moderately increase the model capacity by adopting $(16, 8, 5)$. This configuration achieves a better trade-off between runtime and peak memory usage, while ranking first or second across most datasets. A moderate increase in the embedding dimension $C_*$ enhances representation capability without sacrificing efficiency, still reducing parameters by over 99\% compared to UNet (8L, 2Res). Overall, increasing model capacity within the SL framework consistently improves segmentation performance, aligning with general deep learning principles.

\textbf{Efficiency Analysis Across Training and Testing.} \textbf{During Training Phase}, SL-UNet exhibits increased per-image processing time compared to UNet for certain configurations, particularly as the embedding dimension or number of top selected slices increases. This is mainly due to the increased computational cost introduced by spatial-lifting, which expands 2D feature representations into a pseudo-3D volume and thus incurs additional 3D convolutional operations. Moreover, the limited support of existing CUDA kernels for such operations (e.g., spatially parallel shifting) further reduces computational efficiency in practice. Although, in theory, models with fewer parameters and lower FLOPs should achieve faster test, the current hardware-level implementation constraints prevent full realization of this advantage. Despite the increased training time, these operations help reduce peak memory usage. Moreover, \textbf{During Testing Phase}, the per-image processing time shows negligible increase, while peak memory usage is substantially reduced. These two characteristics, reduced training peak memory and efficient test, make SL-UNet advantageous for large-scale or memory-constrained tasks.

% \textbf{During test with model reliability evaluation}, the per-image processing time of SL-UNet is slightly higher than that of the baseline UNet. For example, selecting the Top-5 slices increases test time to roughly five times that of a single forward pass, which is comparable to performing five stochastic forward passes with MC Dropout and is therefore acceptable. Interestingly, when increasing the number of selected slices to Top-8 (i.e., configuration $(16,8,8)$), the total test time remains significantly lower than performing eight stochastic forward passes with MC Dropout. This demonstrates that the SL architecture has strong potential to reduce uncertainty estimation overhead, and the advantage becomes more pronounced as $s$ increases, since repeated full forward passes are avoided. 

\begin{table}[t]
\footnotesize
\centering
\caption{Comparison of UNet and SL-UNet across different configurations $(m, C_{*}, s)$. The \textbf{best} results are highlighted in \textbf{bold and underlined}, while the second-best results are \textbf{underlined only}.}

\resizebox{\textwidth}{!}{
\begin{tabular}{lcccccc}
\toprule
\multirow{3}{*}{Metric} 
& \multirow{2}{*}{UNet (8L, 2Res)} 
& \multicolumn{5}{c}{SL-UNet (8L, 2Res)} \\
\cmidrule(lr){3-7}
& 
& (16, 4, 5) 
& (16, 8, 5) 
& (16, 16, 5) 
& (32, 8, 5) 
& (16, 8, 8) \\
\midrule
Params (M) & 43.2355 & 0.0121 & 0.0464 & 0.1815 & 0.0121 & 0.0464 \\
MACs (G) & 20.7387 & 2.4994 & 7.0713 & 22.8098 & 4.9989 & 7.0713 \\
\midrule
\multicolumn{7}{c}{\textbf{Training Phase}} \\
\midrule
Time / image (s) & 0.0897 & 0.1161 & 0.1302 & 0.1558 & 0.2794 & 0.1302 \\
Peak GPU Mem (MB) & 866.8 & 445.0 & 634.4 & 1033.6 & 1256.1 & 634.4 \\
\midrule
% \multicolumn{7}{c}{\textbf{Testing Phase}} \\
% \midrule
% Time / image (s) & 0.0353 & 0.1708 & 0.1733 & 0.1755 & 0.5465 & 0.1756 \\
% Peak GPU Mem (MB) & 724.5 & 118.4 & 138.9 & 223.9 & 269.9 & 138.9 \\
% \midrule
\multicolumn{7}{c}{\textbf{Testing Phase}} \\
\midrule
Time / image (s) & 0.0353 & 0.0387 & 0.0398 & 0.0424 & 0.0551 & 0.0398 \\
Peak GPU Mem (MB) & 724.5 & 118.4 & 138.9 & 223.9 & 269.9 & 138.9 \\
\midrule
\multicolumn{7}{c}{\textbf{Datasets (Dice \% $\uparrow$)}} \\
\midrule
DSB2018 & 87.85 & 87.91 & \textbf{\underline{88.91}} & \underline{88.54} & 87.92 & 88.01 \\
Kvasir-SEG & \textbf{\underline{81.16}} & 65.41 & 76.03 & \underline{79.32} & 68.68 & 72.95 \\
ISIC2018 & 82.26 & 82.15 & \textbf{\underline{82.97}} & \underline{82.49} & 76.76 & 77.86 \\
Sartorius & 70.94 & 67.33 & \textbf{\underline{73.02}} & \underline{72.02} & 66.24 & 67.88 \\
MoNuSAC & 69.72 & \underline{70.23} & \textbf{\underline{70.95}} & 70.11 & 66.09 & 67.62 \\
CHASE\_DB1 & \textbf{\underline{73.19}} & 67.48 & \underline{72.39} & 72.28 & 70.81 & 71.22 \\
MoNuSeg & 75.07 & 77.81 & \textbf{\underline{79.02}} & \underline{78.72} & 74.65 & 73.89 \\
TNBC & \textbf{\underline{72.02}} & 67.07 & 69.69 & \underline{70.69} & 63.73 & 62.61 \\
FNC & 61.35 & \underline{67.86} & \textbf{\underline{68.83}} & 66.28 & 60.76 & 61.83 \\
GlaS & 81.65 & 78.09 & \textbf{\underline{84.61}} & \underline{82.63} & 81.08 & 81.36 \\
BCCD & 95.05 & 95.65 & \underline{95.70} & \textbf{\underline{95.79}} & 94.89 & 94.93 \\
FIVES & 78.59 & 76.05 & \underline{79.05} & \textbf{\underline{81.05}} & 78.64 & 78.96 \\
NuInsSeg & \underline{64.59} & 59.42 & \textbf{\underline{65.96}} & 63.73 & 62.19 & 63.74 \\
\bottomrule
\end{tabular}
\label{Tab:different_configurations}
}
\end{table}

\section{Comparison with Lightweight Segmentation Strategies}

In this section, we select the SL-UNet configuration with 8 layers, 2 residual blocks, and hyperparameters $(m=16, C_*=8, s=5)$. To evaluate the advantages of our Spatial Lifting (SL) architecture, we compare it against two representative lightweight segmentation strategies: PixelShuffle (channel-efficient upsampling) and Low-Rank Tensor Factorization (low-rank compression). The chosen SL-UNet variant was selected based on empirical performance, as shown in Table~\ref{Tab:different_configurations}.

\textbf{Alternative Lightweight Methods.} 
For PixelShuffle models, compact variants with low parameter counts and computational cost generally underperform compared to SL-UNet, indicating that naive reduction of parameters and computation leads to insufficient feature representation. Increasing model capacity improves performance on some datasets, even when their parameter counts and computational cost exceed those of our method, but still fails to surpass SL-UNet. On DSB2018, Kvasir-SEG, and ISIC2018, performance significantly declines, likely due to representational degradation caused by excessive network depth that hinders effective feature extraction.

Low-Rank Tensor Factorization models exhibit a similar trend. Even parameter-matched or larger low-rank variants fail to consistently outperform SL-UNet across most datasets, highlighting the superiority of our architectural design over merely increasing parameters or computation.

\begin{table}[t]
\footnotesize
\centering
\caption{Comparison of lightweight segmentation strategies, including PixelShuffle (channel-efficient upsampling) and Low-Rank Tensor Factorization (low-rank compression). SL-UNet is configured with ${(m = 16, C_{*} = 8, s = 5)}$.}

\begin{tabular}{lccccccc}
\toprule
Method & SL-UNet & \multicolumn{2}{c}{PixelShuffle} & \multicolumn{3}{c}{Low-Rank Tensor Factorization} \\
\midrule

Channel 
& {\tiny (8L,2Res)}
& {\tiny [32,64,128,256]} 
& {\tiny [32,64,128,256,512,512]} 
& {\tiny [4,8,16,32,64]} 
& {\tiny [16,32,64,128,256]} 
& {\tiny [32,64,128,256,512]} \\

\midrule
Params (M) & 0.0464 & 1.2425 & 14.2872 & 0.0487 & 0.6941 & 2.7391 \\
MACs (G)   & 7.0713 & 13.8674 & 67.2861 & 0.7835 & 5.9089 & 22.1526 \\

\midrule
Train Time(s)  & 0.1302 & 0.0339 & 0.0857 & 0.1285 & 0.1328 & 0.2062 \\
Train Mem(MB)  & 634.4 & 338.4 & 612.2 & 887.5 & 2313.2 & 4611.9 \\

% \midrule
% Test Time(s)   & 0.1733 & 0.0311 & 0.0351 & 0.0373 & 0.0381 & 0.0468 \\
% Test Mem(MB)   & 138.9 & 131.7 & 332.3 & 48.1 & 131.7 & 275.2 \\

\midrule
Test Time(s) & 0.0398 & 0.0311 & 0.0351 & 0.0373 & 0.0381 & 0.0468 \\
Test Time(MB) & 138.9 & 131.7 & 332.3 & 48.1 & 131.7 & 275.2 \\

\midrule
DSB2018    & \textbf{\underline{88.91}} & 87.51 & 3.59 & 77.55 & 79.31 & 79.96 \\
Kvasir-SEG & \textbf{\underline{76.03}} & 54.89 & 6.86 & 69.56 & 66.64 & 67.83 \\
ISIC2018   & \textbf{\underline{82.97}} & 65.28 & 29.51 & 69.88 & 71.73 & 72.35 \\
Sartorius  & \textbf{\underline{73.02}} & 71.16 & 71.11 & 58.78 & 60.63 & 62.09 \\
MoNuSAC    & \textbf{\underline{70.95}} & 64.98 & 65.85 & 58.95 & 59.49 & 60.64 \\
CHASE\_DB1 & 72.39 & 67.91 & 70.31 & 69.75 & 73.25 & \textbf{\underline{74.09}} \\
MoNuSeg    & \textbf{\underline{79.02}} & 71.06 & 74.93 & 71.71 & 75.11 & 76.97 \\
TNBC       & 69.69 & 66.98 & \textbf{\underline{72.81}} & 59.03 & 68.26 & 66.45 \\
FNC        & \textbf{\underline{68.83}} & 54.95 & 54.62 & 47.27 & 47.64 & 47.99 \\
GlaS       & \textbf{\underline{84.61}} & 71.52 & 80.74 & 78.66 & 80.18 & 80.93 \\
BCCD       & \textbf{\underline{95.70}} & 95.01 & 94.61 & 94.32 & 94.53 & 94.38 \\
FIVES      & \textbf{\underline{79.05}} & 77.79 & 78.24 & 77.15 & 78.75 & 78.88 \\
NuInsSeg   & 65.96 & 64.22 & 64.73 & 66.04 & \textbf{\underline{66.17}} & 65.76 \\

\bottomrule
\end{tabular}
\label{Tab:lightweight}
\end{table}

\textbf{Efficiency and Resource Usage.} 
We further compare per-image training time and memory usage, as summarized in Table~\ref{Tab:lightweight}.  SL-UNet is configured with $(m = 16, C_{*} = 8, s = 5)$, resulting in a compact architecture with $0.0464$M parameters and $7.07$G MACs. Compared with PixelShuffle-based models, SL-UNet requires slightly higher per-image training time ($0.1302$s vs. $0.0339$s and $0.0857$s) and peak training memory (634.4MB vs. 338.4MB and 612.2MB). This is mainly due to PixelShuffle's highly parallel channel rearrangement operation, which reduces intermediate feature-map size and thus achieves lower memory footprint and faster training per image, despite its significantly larger parameter scale (up to 14.29M parameters). In contrast, Low-Rank Tensor Factorization models exhibit consistently higher memory consumption and longer training time. For example, the largest configuration reaches $4611.9$MB memory usage and $0.2062$s per-image training time, significantly exceeding SL-UNet, mainly due to sequential low-rank convolution operations that generate larger intermediate feature representations. In terms of inference efficiency, SL-UNet maintains a low test-time cost ($0.0398$s per image) and moderate memory usage ($138.9$MB), which is comparable to the smaller configuration of PixelShuffle models ($0.0311$s and $131.7$MB). Under the configuration that achieves the best performance, Low-Rank Tensor Factorization methods again show higher inference memory consumption (up to $275.2$MB), reflecting their less efficient feature compression mechanism. Despite PixelShuffle achieving the lowest memory usage and time cost, its segmentation performance is unstable and significantly inferior on several datasets (e.g., $3.59$ on DSB2018 and $6.86$ on Kvasir-SEG in the smallest configuration), indicating that efficient memory usage alone does not guarantee strong representation ability. Overall, SL-UNet achieves a favorable trade-off between efficiency and accuracy. Across 12 datasets, it consistently delivers strong performance, including $88.91$ on DSB2018, $82.97$ on ISIC2018, $95.70$ on BCCD, and $79.05$ on FIVES, while maintaining stable computational overhead.

% Importantly, the SL architecture enables reliable model uncertainty estimation through Top-5 slice selection. Performing the Top-5 slice-based reliability evaluation increases per-image test time to roughly five times that of the non-evaluated version, which is acceptable because it effectively corresponds to five stochastic forward passes with MC Dropout. With parallel computation, the time overhead can be further reduced to less than five times. This demonstrates the practical advantage of our slice-based architecture: it achieves high segmentation performance and provides accessible uncertainty estimation without additional forward computations.

% \textbf{Segmentation Performance.}
% SL-UNet achieves competitive or superior performance on the majority of datasets, including DSB2018, Kvasir-SEG, ISIC2018, MoNuSAC, and GlaS. Notably, these results are obtained with significantly fewer parameters compared to alternative methods, suggesting that the proposed slice-based representation effectively captures task-relevant structural information.

\begin{table}[t]
\centering
\caption{Performance and efficiency comparison between SL-UNet (implicit ensembling) and explicit UNet ensembles with different ensemble sizes.}
\resizebox{0.8\textwidth}{!}{
\begin{tabular}{lcccc}
\toprule
Channel & $[8 * 8]$ & $16 * [8 * 8]$ & $8 * [8 * 8]$ & $1 * [8 * 8]$ \\
\midrule
Model & SL-UNet & Ensemble UNet & Ensemble UNet & Ensemble UNet \\
\midrule
Params (M) & 0.0464 & 0.3738 & 0.1869 & 0.0234 \\
MACs (G) & 7.0713 & 4.8649 & 2.4325 & 0.3041 \\
\midrule
Train Time(s) & 0.1302 & 0.5417 & 0.2649 & 0.0385 \\
Train Mem(MB) & 634.4 & 1040.1 & 526.9 & 78.4 \\
\midrule
Test Time(s) & 0.0398 & 0.0639 & 0.0313 & 0.0051 \\
Test Mem(MB) & 138.9 & 91.3 & 58.8 & 29.3 \\
\midrule
DSB2018 & 88.91 & \underline{\textbf{88.92}} & 88.45 & 87.34 \\
Kvasir-SEG & 76.03 & \underline{\textbf{80.51}} & 79.21 & 74.52 \\
ISIC2018 & 82.97 & \underline{\textbf{87.81}} & 85.54 & 81.45 \\
Sartorius & \underline{\textbf{73.02}} & 71.47 & 71.18 & 69.78 \\
MoNuSAC & \underline{\textbf{70.95}} & 69.55 & 68.66 & 66.01 \\
CHASE\_DB1 & \underline{\textbf{72.39}} & 71.07 & 71.19 & 68.32 \\
MoNuSeg & \underline{\textbf{79.02}} & 73.32 & 73.94 & 72.02 \\
TNBC & \underline{\textbf{69.69}} & 66.53 & 66.23 & 64.97 \\
FNC & \underline{\textbf{68.83}} & 64.11 & 62.29 & 51.39 \\
GlaS & 84.61 & \underline{\textbf{86.58}} & 85.64 & 79.64 \\
BCCD & \underline{\textbf{95.71}} & 94.93 & 94.16 & 93.99 \\
FIVES & \underline{\textbf{79.05}} & 78.36 & 78.15 & 77.82 \\
NuInsSeg & \underline{\textbf{65.96}} & 65.19 & 64.71 & 59.81 \\
\bottomrule
\end{tabular}
}
\label{tab:ensemble}
\end{table}

\begin{figure}[t]
  \noindent
  \begin{tabular}{@{} >{\centering\arraybackslash}m{1cm} m{\dimexpr\textwidth-1cm}@{}}
    (a) & \parbox[t]{\dimexpr\textwidth-1cm}{
          \centering
          \includegraphics[width=0.85\textwidth]{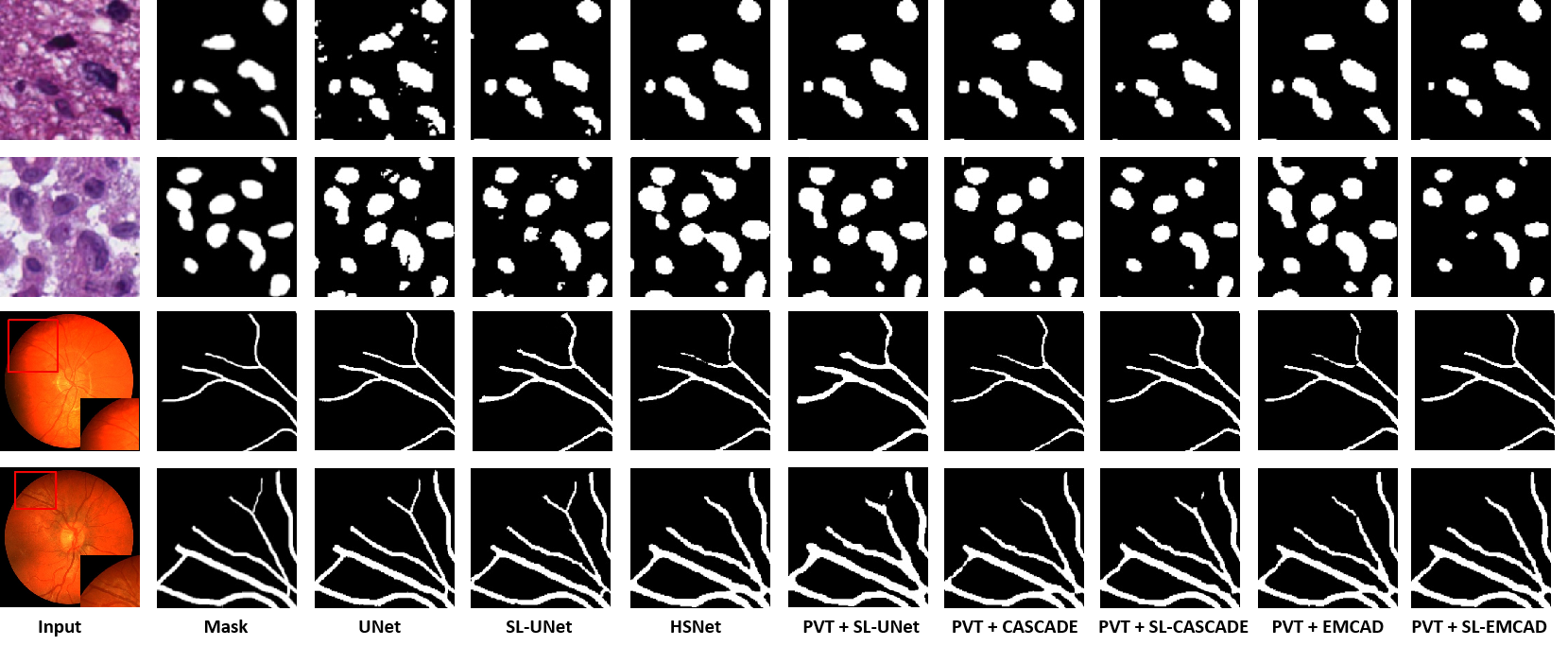}
        } \\
    \noalign{\vskip 0.5cm}
    (b) & \parbox[t]{\dimexpr\textwidth-1cm}{
          \centering
          \includegraphics[width=0.65\textwidth]{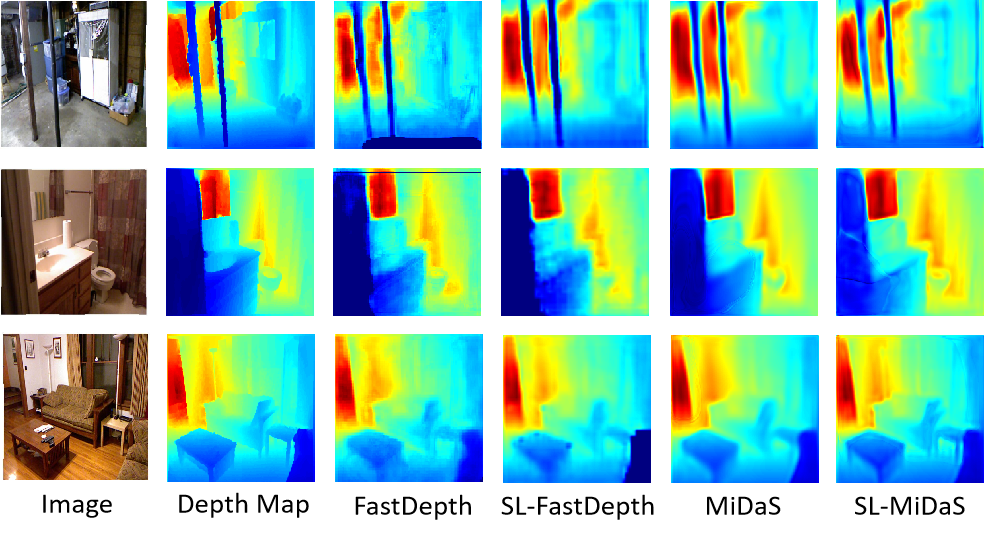}
        } \\
  \end{tabular}
  \caption{(a) Semantic segmentation results; (b) Depth estimation results.}
  \label{fig:rendering}
\end{figure}

% \begin{figure}[t]
%   \centering
%   \includegraphics[width=1\textwidth]{Consistency.png}
%   \caption{Consistency analysis of the proposed quality estimation method. The figure compares the uncertainty estimates generated by our method (Ours, \(T=1\)) against MC Dropout and our own method under different temperature scaling (\(T\)). (a) shows the Pearson correlation (b) shows the Spearman correlation. Asterisk (*) indicates significant correlation (p $<$ 0.05).} 
%   \label{fig:consistency}
% \end{figure}

\section{An Implicit Ensemble-like Representation Framework}
Our proposed SL-UNet addresses segmentation tasks by lifting the input data into a higher-dimensional volumetric representation and processing it with higher-dimensional convolutions. From a conceptual perspective, this design can be interpreted as an \textbf{implicit ensembling framework}, where the additional dimension enables the model to capture diverse and complementary feature representations within a single unified architecture. This stands in contrast to conventional explicit ensembling strategies, which rely on training multiple independent lower-dimensional models and aggregating their outputs. Instead of combining predictions post hoc, SL-UNet integrates feature interactions during representation learning, leading to a more cohesive modeling paradigm. To better understand this perspective, we instantiate this framework in the context of 2D segmentation by lifting images into a pseudo-3D space and compare it with explicit ensembles composed of 1, 8, and 16 narrow-channel ($8 \times 8$) 2D UNet models. 

As shown in Table~\ref{tab:ensemble}, while the 16-model ensemble achieves slightly better performance on several datasets (e.g., DSB2018, Kvasir-SEG, ISIC2018, GlaS), SL-UNet demonstrates more consistent performance across diverse benchmarks, achieving the best results on 9 out of 13 datasets. This suggests that SL-UNet provides improved robustness under dataset variability rather than relying on independent model averaging. From an efficiency perspective, explicit ensembling introduces redundancy due to independently trained models. In contrast, SL-UNet shares parameters within a unified architecture, leading to significantly reduced model size. Specifically, the 16-model ensemble requires 0.3738M parameters, approximately 8× larger than SL-UNet (0.0464M). SL-UNet also reduces training time (0.1302s vs. 0.5417s per image) and memory consumption (634.4 MB vs. 1040.1 MB), demonstrating improved practical efficiency in training and inference. However, SL-UNet incurs higher computational cost in terms of MACs (7.0713G vs. 4.8649G). This reflects a fundamental difference in design philosophy: explicit ensembling improves performance through independent model replication, while SL-UNet increases representational capacity via higher-dimensional feature interaction within a single model. This trade-off highlights a shift from model-level scaling to representation-level scaling, and SL trades higher MACs for fewer parameters and better shared representation.

\section{Comparison with nearly Zero-Cost and Low-Cost Uncertainty Baselines}
In previous sections, we demonstrated that our approach significantly accelerates inference compared to standard Monte Carlo (MC) Dropout while maintaining high uncertainty estimation quality. However, to rigorously validate our claim of achieving robust uncertainty estimation with near-zero overhead, it is imperative to benchmark our approach against established nearly ``zero-cost'' (single forward pass) and low-cost baselines. We comprehensively compare our uncertainty estimation mechanism against three widely utilized baselines: Max-Softmax Confidence, Predictive Entropy and Test-Time Augmentation (TTA) Consistency. To ensure the statistical reliability and generalizability of our findings, we exclude smaller, highly variant datasets and conduct this evaluation across 9 relatively large and robust datasets.

\begin{figure}[t]
    \centering
    \includegraphics[width=1\textwidth]{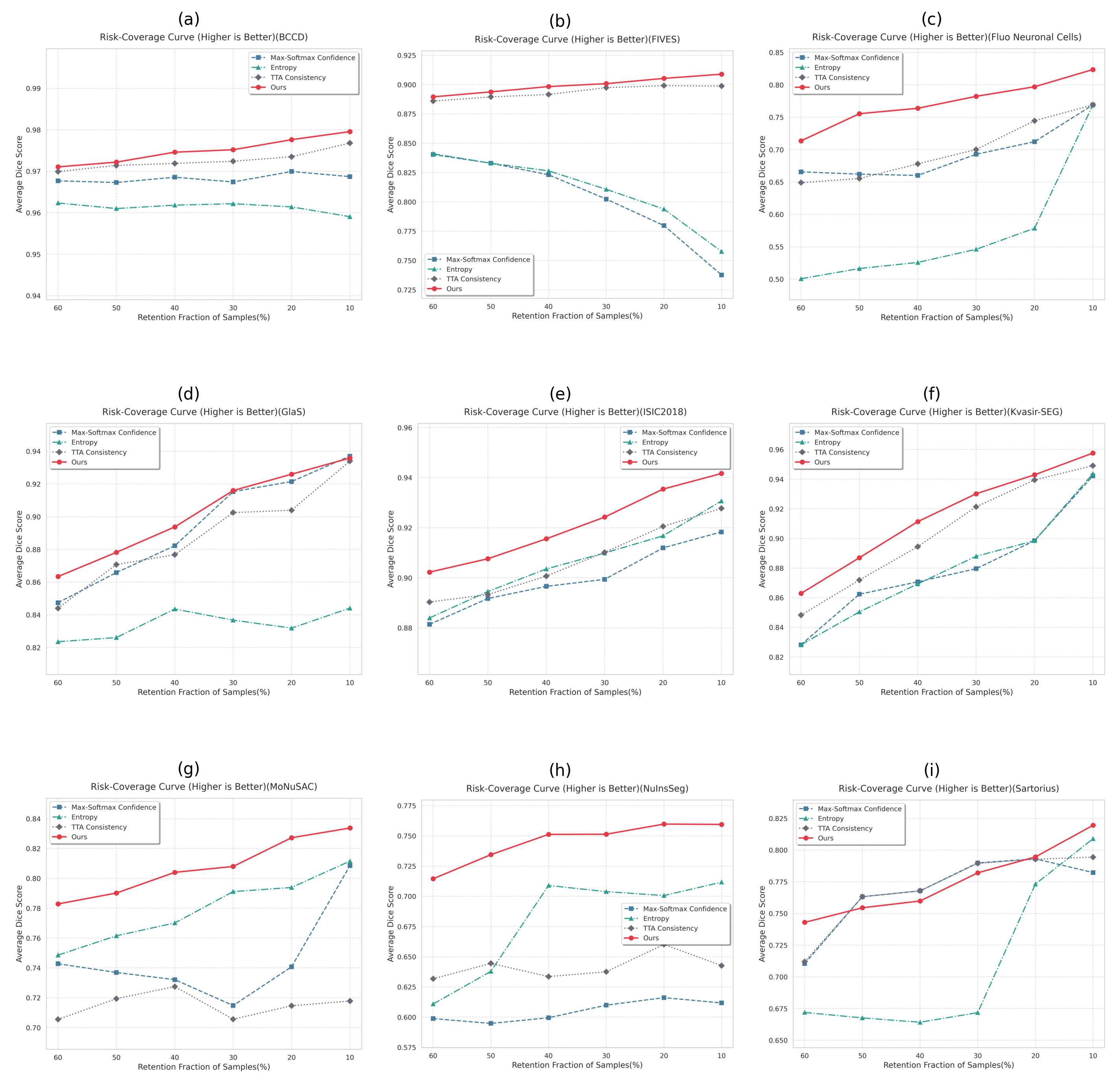} 
    \caption{Risk–coverage curves comparing our proposed single-pass uncertainty estimation method against Max-Softmax Confidence, Predictive Entropy, and TTA Consistency.}
    \label{fig:risk_coverage}
\end{figure}

\textbf{Risk-Coverage Evaluation.} 
We employ Risk-Coverage curves to evaluate the quality of uncertainty estimation, as shown in Fig.~\ref{fig:risk_coverage}. In these curves, the x-axis represents the retention fraction of samples based on the estimated uncertainty (decreasing from 60\% to 10\%), and the y-axis represents the Average Dice Score of the retained subset. An effective uncertainty metric should assign high uncertainty to erroneous predictions. Consequently, as we systematically discard the most ``uncertain'' samples (moving right on the x-axis), the Average Dice Score of the remaining ``certain'' samples should monotonically increase. 

\textbf{Results and Analysis.}
As shown in Fig.~\ref{fig:risk_coverage}, our method (red solid line) consistently outperforms all baselines across nine datasets. Max-Softmax Confidence and Entropy often show suboptimal calibration; in some cases (e.g., BCCD), removing “uncertain” samples does not improve, and may slightly degrade or plateau, Dice performance. TTA Consistency performs better than these zero-cost baselines in several datasets (e.g., BCCD, FIVES, and FNC), but its performance is less stable, with noticeable fluctuations in cases such as MoNuSAC and NuInsSeg. In addition, it requires multiple forward passes, increasing computational cost. In contrast, our method yields a more stable and nearly monotonic risk–coverage trend across datasets, while entropy-based ranking can exhibit abrupt score variations (e.g., FNC and Sartorius), which is undesirable for reliable uncertainty estimation. These results indicate that our approach provides a more consistent uncertainty signal from a single forward pass, outperforming both nearly zero-cost and low-cost baselines.

\section{Visualization Results}
Figure~\ref{fig:rendering} shows qualitative results of our method on two representative visual task: (a) semantic segmentation and (b) depth estimation. 

\section{Analysis of Uncertainty Estimation Effectiveness.}
We evaluate the effectiveness of the proposed uncertainty estimation from two perspectives: (i) agreement with MC Dropout, which requires multiple stochastic forward passes, and (ii) robustness under temperature scaling.

\textbf{Calibration and Efficiency Comparison with MC Dropout.}
We investigate whether the proposed slice-consistency-based measure compromises the model's calibration quality compared to MC Dropout. To this end, we evaluate calibration performance using the Brier score and Expected Calibration Error (ECE), two widely adopted metrics for measuring the accuracy of probabilistic predictions. The Brier score quantifies the mean squared error between predicted probabilities and ground-truth outcomes, while ECE measures the discrepancy between confidence and empirical accuracy. As shown in Table~\ref{tab:correlation}, our method achieves Brier scores and ECE comparable to MC Dropout across multiple datasets, indicating that the calibration quality is well preserved. This finding refutes the concern that the proposed uncertainty estimation degrades the model's original probabilistic behavior. Furthermore, unlike MC Dropout, which requires multiple stochastic forward passes, the proposed method relies on a single forward pass, resulting in a substantially lower test cost. As demonstrated in Table~\ref{tab:correlation}, this leads to an approximate $5\times$ speedup.

\textbf{Robustness under Temperature Scaling.}
Table~\ref{tab:temperature_scaling} validates the robustness of the proposed uncertainty estimation under temperature scaling ($T \in {0.5, 1, 2}$).  In particular, the stability of Brier score and ECE, with standard deviations mostly below 0.01, suggests a calibration behavior that is not driven by extreme or isolated cases, but instead reflects a consistent property of the model’s uncertainty estimation. Consequently, the proposed method is demonstrated to be not an artifact of a specific temperature configuration or extreme-value effects, but a stable and intrinsic behavior that generalizes across calibration settings.

\begin{table}[t]
\footnotesize
\centering
\caption{Comparison of uncertainty estimation across multiple datasets. Brier score and ECE evaluate calibration performance at the dataset level. Time cost per image is reported to assess computational efficiency.}
\begin{tabular*}{\linewidth}{@{\extracolsep{\fill}} ll|cccc @{}}
\toprule
\textbf{Dataset} & \textbf{Method} & \textbf{Brier} & \textbf{ECE} & \textbf{Time Cost (s)} \\
\midrule

\multirow{2}{*}{\textbf{DSB2018}}
& \textbf{Ours}       & 0.021 $\pm$ 0.035  & 0.022 $\pm$ 0.037  & 0.0261  \\
& MC Dropout          & 0.021 $\pm$ 0.034  & 0.021 $\pm$ 0.036  & 0.1447  \\
\midrule

\multirow{2}{*}{\textbf{Kvasir-SEG}}
& \textbf{Ours}       & 0.073 $\pm$ 0.061  & 0.091 $\pm$ 0.066  & 0.0258  \\
& MC Dropout          & 0.081 $\pm$ 0.064  & 0.094 $\pm$ 0.067  & 0.1461  \\
\midrule

\multirow{2}{*}{\textbf{ISIC2018}}
& \textbf{Ours}       & 0.090 $\pm$ 0.139  & 0.119 $\pm$ 0.158  & 0.0185  \\
& MC Dropout          & 0.097 $\pm$ 0.144  & 0.122 $\pm$ 0.163  & 0.1021  \\
\midrule

\multirow{2}{*}{\textbf{Sartorius}}
& \textbf{Ours}       & 0.046 $\pm$ 0.033  & 0.048 $\pm$ 0.042  & 0.0266  \\
& MC Dropout          & 0.045 $\pm$ 0.033  & 0.050 $\pm$ 0.033  & 0.1536  \\
\midrule

\multirow{2}{*}{\textbf{MoNuSAC}}
& \textbf{Ours}       & 0.112 $\pm$ 0.070  & 0.105 $\pm$ 0.079  & 0.0141  \\
& MC Dropout          & 0.111 $\pm$ 0.070  & 0.104 $\pm$ 0.078  & 0.0767  \\
\midrule

\multirow{2}{*}{\textbf{CHASE\_DB1}}
& \textbf{Ours}       & 0.025 $\pm$ 0.003  & 0.023 $\pm$ 0.003  & 0.0308  \\
& MC Dropout          & 0.025 $\pm$ 0.004  & 0.029 $\pm$ 0.004  & 0.1588  \\
\midrule

\multirow{2}{*}{\textbf{MoNuSeg}}
& \textbf{Ours}       & 0.067 $\pm$ 0.014  & 0.082 $\pm$ 0.029  & 0.0277  \\
& MC Dropout          & 0.067 $\pm$ 0.014  & 0.081 $\pm$ 0.028  & 0.1463  \\
\midrule

\multirow{2}{*}{\textbf{TNBC}}
& \textbf{Ours}       & 0.060 $\pm$ 0.029  & 0.053 $\pm$ 0.025  & 0.0105  \\
& MC Dropout          & 0.059 $\pm$ 0.030  & 0.052 $\pm$ 0.025  & 0.0565  \\
\midrule

\multirow{2}{*}{\textbf{FNC}}
& \textbf{Ours}       & 0.025 $\pm$ 0.006  & 0.025 $\pm$ 0.006  & 0.0132  \\
& MC Dropout          & 0.025 $\pm$ 0.006  & 0.026 $\pm$ 0.007  & 0.0719  \\
\midrule

\multirow{2}{*}{\textbf{GlaS}}
& \textbf{Ours}       & 0.153 $\pm$ 0.079  & 0.143 $\pm$ 0.104  & 0.0271  \\
& MC Dropout          & 0.160 $\pm$ 0.074  & 0.144 $\pm$ 0.103  & 0.1541  \\
\midrule

\multirow{2}{*}{\textbf{BCCD}}
& \textbf{Ours}       & 0.014 $\pm$ 0.012  & 0.015 $\pm$ 0.014  & 0.0099  \\
& MC Dropout          & 0.015 $\pm$ 0.013  & 0.016 $\pm$ 0.016  & 0.0525  \\
\midrule

\multirow{2}{*}{\textbf{FIVES}}
& \textbf{Ours}       & 0.019 $\pm$ 0.011  & 0.025 $\pm$ 0.014  & 0.0228  \\
& MC Dropout          & 0.020 $\pm$ 0.010  & 0.024 $\pm$ 0.013  & 0.1255  \\
\midrule

\multirow{2}{*}{\textbf{NuInsSeg}}
& \textbf{Ours}       & 0.087 $\pm$ 0.061  & 0.087 $\pm$ 0.061  & 0.0196  \\
& MC Dropout          & 0.084 $\pm$ 0.063  & 0.096 $\pm$ 0.077  & 0.1131  \\

\bottomrule
\end{tabular*}
\label{tab:correlation}
\end{table}

\begin{table}[t]
\footnotesize
\centering
\caption{
Comparison of uncertainty estimation methods with actual Dice scores. Each method reports Pearson (\(r\)) and Spearman (\(\rho\)) correlation with actual Dice scores. Ours (\(T\)) refers to temperature scaling. Calibration errors (Brier / ECE) are reported as mean ± standard deviation across \(T \in \{0.5, 1, 2\}\). Asterisk (*) indicates significant correlation (p $<$ 0.05).
}
\resizebox{1\linewidth}{!}{
\begin{tabular}{l|cc cc cc cc}
\toprule
Dataset 
& \multicolumn{2}{c}{Ours (\(T=1\))} 
& \multicolumn{2}{c}{Ours (\(T=0.5\))} 
& \multicolumn{2}{c}{Ours (\(T=2\))} 
& \multicolumn{2}{c}{Calibration} \\
\cmidrule(lr){2-3} \cmidrule(lr){4-5} \cmidrule(lr){6-7} \cmidrule(lr){8-9}
& r & \(\rho\) & r & \(\rho\) & r & \(\rho\) & Brier & ECE \\
\midrule

DSB2018*    & 0.68 & 0.66 & 0.62 & 0.80 & 0.64 & 0.84 & $0.035 \pm 0.001$ & $0.040 \pm 0.001$ \\
Kvasir-SEG* & 0.67 & 0.65 & 0.71 & 0.70 & 0.73 & 0.77 & $0.132 \pm 0.002$ & $0.133 \pm 0.002$ \\
ISIC2018*   & 0.80 & 0.73 & 0.73 & 0.67 & 0.72 & 0.67 & $0.093 \pm 0.003$ & $0.095 \pm 0.002$ \\
Sartorius*  & 0.79 & 0.77 & 0.72 & 0.78 & 0.73 & 0.79 & $0.072 \pm 0.002$ & $0.070 \pm 0.002$ \\
MoNuSAC*    & 0.73 & 0.56 & 0.77 & 0.60 & 0.77 & 0.62 & $0.182 \pm 0.008$ & $0.180 \pm 0.011$ \\
CHASE\_DB1   & 0.67 & 0.57 & 0.72 & 0.72 & 0.69 & 0.54 & $0.021 \pm 0.005$ & $0.026 \pm 0.002$ \\
MoNuSeg*    & 0.55 & 0.41 & 0.65 & 0.47 & 0.59 & 0.48 & $0.079 \pm 0.010$ & $0.095 \pm 0.008$ \\
TNBC*       & 0.65 & 0.68 & 0.55 & 0.67 & 0.71 & 0.76 & $0.091 \pm 0.006$ & $0.089 \pm 0.008$ \\
FNC*        & 0.68 & 0.77 & 0.52 & 0.78 & 0.40 & 0.69 & $0.031 \pm 0.001$ & $0.045 \pm 0.019$ \\
GlaS*       & 0.64 & 0.67 & 0.51 & 0.63 & 0.50 & 0.61 & $0.210 \pm 0.011$ & $0.219 \pm 0.014$ \\
BCCD*       & 0.83 & 0.67 & 0.87 & 0.75 & 0.86 & 0.76 & $0.022 \pm 0.002$ & $0.029 \pm 0.003$ \\
FIVES*      & 0.88 & 0.84 & 0.88 & 0.88 & 0.85 & 0.87 & $0.031 \pm 0.002$ & $0.038 \pm 0.004$ \\
NuInsSeg*   & 0.70 & 0.72 & 0.69 & 0.70 & 0.72 & 0.72 & $0.098 \pm 0.004$ & $0.098 \pm 0.002$ \\
\bottomrule
\end{tabular}
}
\label{tab:temperature_scaling}
\end{table}

\section{Additional Theoretical Analysis of Spatial Lifting}
\label{app:additional_theory}

This section provides a more detailed theoretical interpretation of Spatial Lifting (SL). 
The goal is not to claim that lifting injects new information into the input, since the lifted
volume is initialized by replication. Instead, we show that SL can be understood as a finite-axis,
boundary-conditioned family of coupled predictors, trained with an implicit consensus regularizer
and aggregated through a product-of-experts-like rule.

\subsection{Finite-axis lifting as a coupled family of 2D predictors}
\label{app:finite_axis_family}

Let the lifted input be \(X_z=x\) for \(z=1,\ldots,m\), where \(x\) is the original 2D input.
Consider a stack of 3D convolutional layers whose kernels are local along the lifted axis. For
simplicity, write the \(z\)-axis radius of layer \(\ell\) as \(r_\ell\). The feature at slice \(z\)
after layer \(\ell+1\) has the form
\begin{equation}
h^{(\ell+1)}_z
=
\psi^{(\ell)}
\left(
\sum_{\tau=-r_\ell}^{r_\ell}
\mathbf{1}\{1\le z-\tau\le m\}
K^{(\ell)}_{\tau} *_{xy} h^{(\ell)}_{z-\tau}
\right),
\end{equation}
where \(*_{xy}\) denotes convolution in the native spatial dimensions. The indicator term is the
only difference between an infinite or circular lifted axis and the finite lifted axis used by SL.

For a single linear lifted convolution applied to the replicated input, the output slice is
\begin{equation}
Y_z
=
\sum_{\tau=-r}^{r}
\mathbf{1}\{1\le z-\tau\le m\}
K_{\tau}*_{xy}x
=
K^{\mathrm{eff}}_z*_{xy}x,
\end{equation}
where
\begin{equation}
K^{\mathrm{eff}}_z
=
\sum_{\tau=-r}^{r}
\mathbf{1}\{1\le z-\tau\le m\}K_{\tau}.
\end{equation}
Thus different slices implement different effective 2D filters only because the auxiliary axis is
finite and non-periodic. In a deeper nonlinear network, the exact linear effective kernel expression
no longer holds, but the same mechanism gives a slice-dependent mapping
\begin{equation}
P_z = g_{\theta,z}(x), \qquad z=1,\ldots,m,
\end{equation}
where all \(g_{\theta,z}\) share parameters and differ through their boundary-conditioned
computational paths.

This observation gives an important sanity check. If the lifted axis used circular padding and no
slice-dependent positional information, then a replicated input would remain identical across slices
under any \(z\)-translation-equivariant network. In that case, SL would collapse to a single predictor
repeated \(m\) times, and slice consistency would no longer provide a useful uncertainty signal.
Therefore, the useful diversity in SL is not caused by replication itself, but by finite-axis local
interaction and boundary-conditioned computation.

\subsection{When do slices become different?}
\label{app:diversity_condition}

Let
\begin{equation}
R_L=\sum_{\ell=1}^{L} r_\ell
\end{equation}
be the one-sided receptive radius along the lifted axis for an \(L\)-layer stride-one stack. If a slice
\(z\) satisfies
\begin{equation}
R_L < z-1
\quad\text{and}\quad
R_L < m-z,
\end{equation}
then its computational neighborhood does not touch either boundary. For a replicated input and a
translation-equivariant stack along \(z\), all such interior slices receive identical lifted-axis context
and are therefore indistinguishable. Non-trivial slice diversity appears only when the effective
\(z\)-axis receptive field reaches the finite boundaries, or when the architecture contains operations
that otherwise break \(z\)-translation equivariance.

This explains the empirical behavior observed in the slice-selection study. Boundary slices can be
degraded because they see truncated context, while intermediate slices often provide a better balance:
they are influenced by the finite-axis structure but are less dominated by padding artifacts. Hence,
the best slices are expected to concentrate in interior regions rather than at the extreme boundaries.

\subsection{Dense slice supervision as an implicit consensus regularizer}
\label{app:bregman_consensus}

At one spatial location, let \(u_1,\ldots,u_m\in\mathbb{R}^C\) denote slice-wise logits and let \(y\)
be the common target. Suppose the per-slice loss can be written as
\begin{equation}
\ell(y,u)=\Phi(u)-\langle T(y),u\rangle+c(y),
\end{equation}
where \(\Phi\) is convex and differentiable. This includes multiclass cross-entropy with
\(\Phi(u)=\log\sum_c\exp(u_c)\), binary logistic loss, and squared loss on logits or probabilities.
Let
\begin{equation}
\bar u = \frac{1}{m}\sum_{z=1}^{m}u_z
\end{equation}
and define the Bregman divergence
\begin{equation}
D_{\Phi}(a,b)
=
\Phi(a)-\Phi(b)-\langle \nabla\Phi(b),a-b\rangle .
\end{equation}
Then the dense slice-supervised objective satisfies the exact identity
\begin{equation}
\frac{1}{m}\sum_{z=1}^{m}\ell(y,u_z)
=
\ell(y,\bar u)
+
\frac{1}{m}\sum_{z=1}^{m}D_{\Phi}(u_z,\bar u).
\end{equation}
The first term trains the mean predictor, while the second term penalizes disagreement among the
slice-wise logits in the geometry induced by \(\Phi\). For softmax cross-entropy,
\begin{equation}
D_{\Phi}(u_z,\bar u)
=
\mathrm{KL}\!\left(
\mathrm{softmax}(\bar u)\,\|\,\mathrm{softmax}(u_z)
\right),
\end{equation}
so the consensus term acts directly on the predictive distributions. It does not penalize
class-independent logit shifts, which are irrelevant to softmax predictions.

If \(\Phi\) is \(\mu\)-strongly convex on the considered subspace, then
\begin{equation}
\frac{1}{m}\sum_{z=1}^{m}D_{\Phi}(u_z,\bar u)
\ge
\frac{\mu}{2m}\sum_{z=1}^{m}\|u_z-\bar u\|_2^2
=
\frac{\mu}{2m^2}\sum_{1\le i<j\le m}\|u_i-u_j\|_2^2 .
\end{equation}
Thus dense slice supervision can be viewed as optimizing the average predictor while implicitly
discouraging pairwise slice disagreement.

For Dice or IoU-style losses, the above Bregman identity is not exact. However, when the loss is
smooth in a local neighborhood of \(\bar u\), a second-order expansion gives
\begin{equation}
\frac{1}{m}\sum_{z=1}^{m}\ell(y,u_z)
=
\ell(y,\bar u)
+
\frac{1}{2m}\sum_{z=1}^{m}
(u_z-\bar u)^\top
\nabla_u^2\ell(y,\bar u)
(u_z-\bar u)
+
O\!\left(\max_z\|u_z-\bar u\|_2^3\right).
\end{equation}
Therefore, for non-Bregman losses, dense slice supervision should be interpreted as a local
consensus effect rather than an exact global decomposition.

\subsection{Logit aggregation as product-of-experts consensus}
\label{app:poe_aggregation}

The testing rule in SL aggregates selected slice logits. Let \(Z\) be the selected slice set and let
\(u_z(c)\) be the logit for class \(c\) at slice \(z\). The aggregated prediction is
\begin{equation}
\hat q_Z(c)
=
\mathrm{softmax}_c\left(\sum_{z\in Z}u_z\right).
\end{equation}
Let \(q_z(c)=\mathrm{softmax}_c(u_z)\). Since
\begin{equation}
q_z(c)=\frac{\exp(u_z(c))}{\sum_d\exp(u_z(d))},
\end{equation}
we have
\begin{equation}
\hat q_Z(c)
=
\frac{\prod_{z\in Z}q_z(c)}
{\sum_d \prod_{z\in Z}q_z(d)} .
\end{equation}
Thus summing logits is equivalent to a normalized product of slice-wise predictive distributions.
This is a product-of-experts consensus rule: a class receives high probability only when it is supported
by multiple selected slices. If average logits are used instead,
\begin{equation}
\tilde q_Z(c)
=
\mathrm{softmax}_c\left(\frac{1}{|Z|}\sum_{z\in Z}u_z\right)
=
\frac{\left(\prod_{z\in Z}q_z(c)\right)^{1/|Z|}}
{\sum_d \left(\prod_{z\in Z}q_z(d)\right)^{1/|Z|}},
\end{equation}
which corresponds to a normalized geometric mean of slice probabilities. The sum-logit rule is
sharper, while the average-logit rule is usually better calibrated. This provides a theoretical reason
why temperature scaling can be useful when evaluating uncertainty and calibration.

For binary segmentation, the same rule can be written in terms of odds:
\begin{equation}
\frac{\hat q_Z}{1-\hat q_Z}
=
\prod_{z\in Z}
\frac{q_z}{1-q_z}.
\end{equation}
Hence logit aggregation combines slice evidence multiplicatively.

\subsection{Why slice consistency can predict segmentation quality}
\label{app:quality_bound}

Let \(p_z\) be the slice-wise probability map after activation. Define the selected-set average and
unselected-set average as
\begin{equation}
\bar p_Z=\frac{1}{|Z|}\sum_{i\in Z}p_i,
\qquad
\bar p_{Z^c}=\frac{1}{|Z^c|}\sum_{j\in Z^c}p_j.
\end{equation}
Define the cross-set dispersion
\begin{equation}
\Delta_Z
=
\frac{1}{|Z||Z^c|}
\sum_{i\in Z}\sum_{j\in Z^c}
\|p_i-p_j\|_2^2 .
\end{equation}
By Jensen's inequality,
\begin{equation}
\|\bar p_Z-\bar p_{Z^c}\|_2^2
\le
\Delta_Z .
\end{equation}
Therefore, for any \(L\)-Lipschitz evaluation loss \(\mathcal{E}(y,p)\),
\begin{equation}
\left|
\mathcal{E}(y,\bar p_Z)
-
\mathcal{E}(y,\bar p_{Z^c})
\right|
\le
L\|\bar p_Z-\bar p_{Z^c}\|_2
\le
L\sqrt{\Delta_Z}.
\end{equation}
This bound shows that low cross-set dispersion makes the selected aggregate and the remaining
slice family behave similarly under any Lipschitz quality measure. Since all slices are trained toward
the same ground-truth target, this provides a theoretical basis for using cross-slice agreement as a
quality proxy.

The practical score \(Q\) uses pairwise Dice agreement between binarized selected and unselected
predictions:
\begin{equation}
Q
=
\frac{1}{|Z||Z^c|}
\sum_{i\in Z}\sum_{j\in Z^c}
\mathrm{Dice}(B_i,B_j).
\end{equation}
For binary masks \(B_i,B_j\),
\begin{equation}
1-\mathrm{Dice}(B_i,B_j)
=
\frac{|B_i\triangle B_j|}{|B_i|+|B_j|},
\end{equation}
where \(B_i\triangle B_j\) is the symmetric difference. Thus high \(Q\) corresponds to low normalized
mask disagreement, which is a task-specific surrogate for low cross-set dispersion. This also clarifies
a limitation: if all slices agree on the same wrong prediction, \(Q\) can still be high. Therefore,
\(Q\) measures self-consistency rather than ground-truth accuracy directly.

A natural uncertainty score is therefore
\begin{equation}
U_{\mathrm{disp}} = 1-Q,
\end{equation}
while the variance of pairwise Dice similarities can be reported as an auxiliary heterogeneity score:
\begin{equation}
U_{\mathrm{var}}
=
\mathrm{Var}_{i\in Z,\,j\in Z^c}
\left[
\mathrm{Dice}(B_i,B_j)
\right].
\end{equation}
In practice, \(U_{\mathrm{disp}}\) is more directly aligned with risk ranking, whereas
\(U_{\mathrm{var}}\) describes how uneven the cross-slice agreement pattern is.

\subsection{Bias--variance view of top-\texorpdfstring{\(s\)}{s} slice aggregation}
\label{app:bias_variance_top_s}

Assume each selected slice prediction can be decomposed as
\begin{equation}
p_z = p^\star + b_z + \varepsilon_z,
\qquad
\mathbb{E}[\varepsilon_z]=0,
\end{equation}
where \(p^\star\) is the desired prediction, \(b_z\) is a slice-dependent bias, and \(\varepsilon_z\)
is a residual error. The selected average satisfies
\begin{equation}
\mathbb{E}\|\bar p_Z-p^\star\|_2^2
=
\left\|
\frac{1}{|Z|}\sum_{z\in Z}b_z
\right\|_2^2
+
\mathrm{tr}
\left(
\frac{1}{|Z|^2}
\sum_{i,j\in Z}
\mathrm{Cov}(\varepsilon_i,\varepsilon_j)
\right).
\end{equation}
If the selected slices have identical residual variance \(\sigma^2\) and average pairwise correlation
\(\rho\), the variance term scales as
\begin{equation}
\sigma^2\frac{1+(|Z|-1)\rho}{|Z|}.
\end{equation}
Aggregation is most beneficial when residual errors are not perfectly correlated. If \(\rho=1\),
averaging gives no variance reduction; if \(\rho=0\), the variance decreases approximately as
\(1/|Z|\). However, increasing \(|Z|\) can also introduce more biased slices. This explains why
selecting a small set of reliable slices can outperform using all slices: top-\(s\) selection balances
bias reduction and variance reduction.

\subsection{Efficiency condition for low-width lifted computation}
\label{app:efficiency_condition}

A 3D convolution is not automatically cheaper than a 2D convolution. The efficiency of SL comes
from combining lifting with a narrow lifted representation. Consider a 2D layer with spatial area
\(A=HW\), kernel size \(k_xk_y\), and channels \(C_{\ell}\rightarrow C_{\ell+1}\). Its MACs scale as
\begin{equation}
\mathrm{MAC}_{2D}
\propto
A\,k_xk_y\,C_{\ell}C_{\ell+1}.
\end{equation}
A lifted 3D layer with lifted depth \(m\), depth kernel size \(k_z\), and fixed width \(C^\ast\) has
\begin{equation}
\mathrm{MAC}_{SL}
\propto
A\,m\,k_xk_yk_z\,(C^\ast)^2.
\end{equation}
Thus the per-layer MAC ratio is approximately
\begin{equation}
\frac{\mathrm{MAC}_{SL}}{\mathrm{MAC}_{2D}}
\approx
\frac{m\,k_z\,(C^\ast)^2}{C_{\ell}C_{\ell+1}}.
\end{equation}
Similarly, the parameter ratio is approximately
\begin{equation}
\frac{\mathrm{Param}_{SL}}{\mathrm{Param}_{2D}}
\approx
\frac{k_z\,(C^\ast)^2}{C_{\ell}C_{\ell+1}}.
\end{equation}
Therefore, SL is computationally favorable when
\begin{equation}
C^\ast
<
\sqrt{\frac{C_{\ell}C_{\ell+1}}{m\,k_z}}
\end{equation}
for MAC reduction, and when
\begin{equation}
C^\ast
<
\sqrt{\frac{C_{\ell}C_{\ell+1}}{k_z}}
\end{equation}
for parameter reduction. This condition makes clear that the reported efficiency gains should be
interpreted as the result of finite-axis lifting together with narrow-channel lifted processing, rather
than lifting alone.

The number of selected slices \(s\) affects only the final aggregation cost if all \(m\) slices are
computed in a single forward pass. Hence \(s\) should have negligible influence on network MACs
unless the implementation recomputes or separately processes selected slices.

\subsection{Limitations predicted by the theory}
\label{app:theory_limitations}

The above analysis suggests several failure modes and testable predictions.

First, if circular padding is used along the lifted axis, or if the finite-axis boundary effect is removed,
the slice family should collapse toward identical predictions. In that case, SL loses both ensemble-like
diversity and self-consistency-based uncertainty estimation.

Second, if the lifted depth \(m\) is much larger than the effective \(z\)-axis receptive field, many
interior slices may become redundant. Conversely, if \(m\) is too small, all slices may be dominated
by boundary artifacts. The best regime is expected to occur when the lifted-axis receptive field is
large enough to create meaningful slice-dependent predictors but not so small that only boundary
slices differ.

Third, high slice consistency does not guarantee correctness. When all slice predictors share the
same systematic bias, \(Q\) may be high even for an inaccurate prediction. This is a general limitation
of self-consistency-based quality estimation. Combining \(Q\) with confidence calibration or
out-of-distribution detection can further improve reliability.

\section{Limitations and Future Work}
\label{sec:limitations}
While the proposed Spatial Lifting (SL) framework demonstrates strong performance and efficiency across a range of tasks, several limitations remain. First, SL employs a fixed expansion into a higher-dimensional representation (e.g., depth $m=16$), which may not be optimal for all datasets and tasks; in principle, the lifting dimensionality should be adaptive, and automated approaches such as neural architecture search or meta-learning could be explored to determine more suitable configurations. In addition, the current implementation relies on standard convolutional architectures (e.g., 3D U-Net) to process lifted representations, which, while convenient, may not fully exploit the structural properties introduced by spatial lifting; extending SL to architectures specifically tailored to lifted representations, as well as to transformer-based models, could further enhance both representational capacity and computational efficiency. Furthermore, despite our method substantially reducing parameter count and theoretical computational cost (e.g., GMACs or FLOPs), peak memory usage and wall-clock time do not decrease proportionally, suggesting a gap between theoretical and practical efficiency that we attribute to a mismatch between the SL computation pattern and current hardware/software stacks, particularly the limited support of existing CUDA kernels. Finally, as a self-consistency-based metric, the proposed quality score $Q$ has inherent limitations: high slice consistency does not necessarily imply correctness, and when slice predictors share systematic biases, $Q$ may remain high even for inaccurate predictions; incorporating complementary signals such as confidence calibration or out-of-distribution detection may help improve reliability.

\end{document}